%% file: main.tex
\documentclass[10pt,twocolumn,letterpaper]{article}

\usepackage[pagenumbers]{wacv} 
\usepackage{multirow}

\definecolor{wacvblue}{rgb}{0.21,0.49,0.74}
\usepackage[pagebackref,breaklinks,colorlinks,allcolors=wacvblue]{hyperref}
\usepackage[accsupp]{axessibility}
\def\wacvPaperID{1625} 
\def\confName{WACV}
\def\confYear{2027}

\title{Intra-Architecture Source Attribution for Stable Diffusion:\\ A New Task with Benchmark Dataset and Baseline}

\author{Ashutosh Dayal, Mallika Garg, and Debashis Sen\\
Indian Institute of Technology Kharagpur, INDIA\\
{\tt\small ashutoshdayal@kgpian.iitkgp.ac.in, mallika@ec.iitr.ac.in, dsen@ece.iitkgp.ac.in}
}

\begin{document}
\maketitle
\input{sec/0_abstract}    
\input{sec/1_intro}
\input{sec/2_formatting}
\input{sec/3_finalcopy}
\vspace{-2mm}
{\vspace{-2mm}
    \small
    \bibliographystyle{ieeenat_fullname}
    \bibliography{main}
}

\end{document}


\maketitle

We provide the following important details about our contribution in this supplementary document:
\begin{itemize}
    \item Continuing from Section~3.2 in the \textit{main paper}, we describe a few more \textit{IntrAA-SD} benchmark (ISDB) dataset In-domain (ID) and out-of-distribution (OOD) prompt categories in Section \ref{sec:prompts} here.
    \item Continuing from Section~4.1 in the \textit{main paper}, we provide a visualization to analyze the style feature extraction using CLIP in Section~\ref{sec:clip} here.
    \item Continuing from Section~5.2 in the \textit{main paper}, in Section~\ref{sec:ExistingAttribution} here, we discuss the intra-architecture source attribution results of some existing architecture-level source attribution methods when trained and tested in our ISDB dataset.
    \item In addition to the above, we dive into more experimental analysis of our \textit{IntrAA-SD} intra-architecture source attribution baseline framework in Section~\ref{sec:experiments} here.
\end{itemize}

\section{Certain Prompt Categories of Our ISDB Dataset} \label{sec:prompts}

\begin{table*}[!t]
\begin{center}
\caption{Intra-architecture source attribution accuracy (\%) on our \textit{ISDB} test set. Models -
M1: SDXL Base, M2: Juggernaut, M3: RealVis, M4: DreamShaper, M5: Animagine, M6: Counterfeit, M7: ZavyChroma, M8: Copax,
M9: Proteus, M10: AAM Anime.  \textbf{$^\ast$ Ours} is the only algorithm (\& the first) in the list which is designed for intra-architecture source attribution. All others listed are existing architecture-level source attribution methods.}
\label{tab:main_results}
\begin{tabular}{cc|cccccccccc|c}
\hline
& \textbf{Method} & M1 & M2 & M3 & M4 & M5 & M6 & M7 & M8 & M9 & M10 & \textbf{Overall} \\
\hline
\multirow{6}{*}{\rotatebox{90}{\textbf{ID}}}
& Spec~\cite{frank2020leveraging}  & 18.13 & 12.96 & 11.86 & 46.45 & 40.83 & 36.51 & 9.31  & 20.49 & 32.01 & 21.32 & 24.99 \\
& UnivFD~\cite{clip_baseline}     & 24.46 & 27.79 & 30.32 & 70.52 & 76.47 & 29.73 & 33.85 & 35.62 & 56.71 & 61.41 & 44.69 \\
& DE-FAKE~\cite{sha2023fake}      & 62.96 & 65.37 & 61.32 & 89.58 & 93.91 & 76.24 & 62.28 & 67.97 & 87.87 & 86.22 & 75.37 \\
& NPR~\cite{tan2023learning}      & 88.27 & 85.53 & 86.78 & 97.97 & 96.56 & 94.03 & 86.51 & 89.81 & 94.42 & 95.08 & 91.50 \\
& CNNDetect~\cite{wang2020}       & 87.26 & 83.93 & 88.96 & 98.32 & 98.40 & 95.38 & 87.01 & 90.27 & 96.34 & 96.28 & 92.21 \\
& \textbf{Ours$^\ast$}                    & \textbf{99.42} & \textbf{99.79} & \textbf{98.67} & \textbf{99.83} & \textbf{99.92} & \textbf{98.29} & \textbf{99.67} & \textbf{97.88} & \textbf{98.58} & \textbf{98.21} & \textbf{99.03} \\
\hline
\multirow{6}{*}{\rotatebox{90}{\textbf{OOD}}}
& Spec~\cite{frank2020leveraging}  & 11.50 & 11.70 & 12.40 & 55.00 & 36.90 & 31.10 & 11.50 & 16.20 & 31.40 & 14.40 & 23.21 \\
& UnivFD~\cite{clip_baseline}     & 20.60 & 22.00 & 26.50 & 78.10 & 74.80 & 25.60 & 27.40 & 32.20 & 61.70 & 59.00 & 42.79 \\
& DE-FAKE~\cite{sha2023fake}      & 41.20 & 34.00 & 34.90 & 78.00 & 83.80 & 36.10 & 31.80 & 45.40 & 68.30 & 65.30 & 51.88 \\
& CNNDetect~\cite{wang2020}       & 70.30 & 47.30 & 61.90 & 94.40 & 92.10 & 76.30 & 45.70 & 66.60 & 86.60 & 78.20 & 71.94 \\
& NPR~\cite{tan2023learning}      & 75.20 & 53.60 & 61.90 & 95.10 & 92.00 & 79.60 & 54.20 & 71.10 & 83.80 & 81.90 & 74.84 \\
& \textbf{Ours$^\ast$}                    & \textbf{91.50} & \textbf{70.50} & \textbf{82.50} & \textbf{99.00} & \textbf{98.00} & \textbf{98.00} & \textbf{74.00} & \textbf{90.00} & \textbf{90.50} & \textbf{91.50} & \textbf{88.55} \\
\hline
\end{tabular}%
\end{center}

\end{table*}

\subsection{In-Domain (ID) Generative Stress Test Prompts}
We have considered $7$ prompt categories in the ID set of our ISDB dataset that are designed to push the fine-tuned diffusion generators to their limits, both during training and inference. As in such situations the generative models are more likely to introduce pronounced generator-specific signatures in the images generated~\cite{corvi2023}, they represent favorable cases for attribution methods. We discuss a few such generative stress test categories below.
\begin{itemize}
    \item \textbf{Texture and Geometric \& Repetitive:} Generating textures or repetitive patterns, which often represent high spatial frequencies, challenges generator decoders, especially their spatial up-sampling layers~\cite{rombach2022high}. Different generator fine-tuning processes alter how the fine-tuned models handle aliasing, causing them to produce distinct frequency-domain model-related signatures within the generated image regions~\cite{frank2020leveraging}.
    \item \textbf{Text \& Signals:} Rendering coherent text or a signal plot often leads to failure for the Stable Diffusion XL (SDXL) base model~\cite{sdxl, saharia2022photorealistic}. Fine-tunes try to overcome this weakness by aggressively introducing structural corrections particularly to improve synthetic text generation. However, such attempts often leave model-specific traces in the generated areas containing characters, symbols or plots.
    \item \textbf{Noise \& Malicious:} Such prompts provide minimal scope for semantic conditioning. So, without a meaningful object to render, the generator usually relies heavily on its underlying structural prior, which leaves its imprint in the generated image that aids source isolation without much interference from any visual content~\cite{wang2023dire}.
\end{itemize}
\subsection{Out-Of-Distribution (OOD) Adversarial Prompts}
We have considered $7$ prompt categories in the OOD set of our ISDB dataset that are adversarial in nature to the fine-tuned generative models. The images thus generated tests the source attribution systems, as they must posses the capability of ignoring the uncharacteristic image style generated to a large extent in favor of model signature.
\begin{itemize}
    \item \textbf{Void Prompts:} Prompts such as ``\emph{Pure white background, nothing else}'' or ``\emph{Complete darkness, empty frame}'' demand generation of images with no content. The models are usually not well trained for such generation, and hence, can produce arbitrarily-styled images with model signature camouflaged in them.  
    \item \textbf{Nested Representations:} Prompts such as ``\emph{A photograph of an oil painting displayed on a modern gallery wall}'' requires the generator to transition abruptly between distinct image types often producing anomalous /unseen distortions at boundaries, where model signature may be contained~\cite{clip_baseline}.
\end{itemize}


\begin{figure}[!t]
    \centering
    \includegraphics[width=0.45\textwidth, page=1]{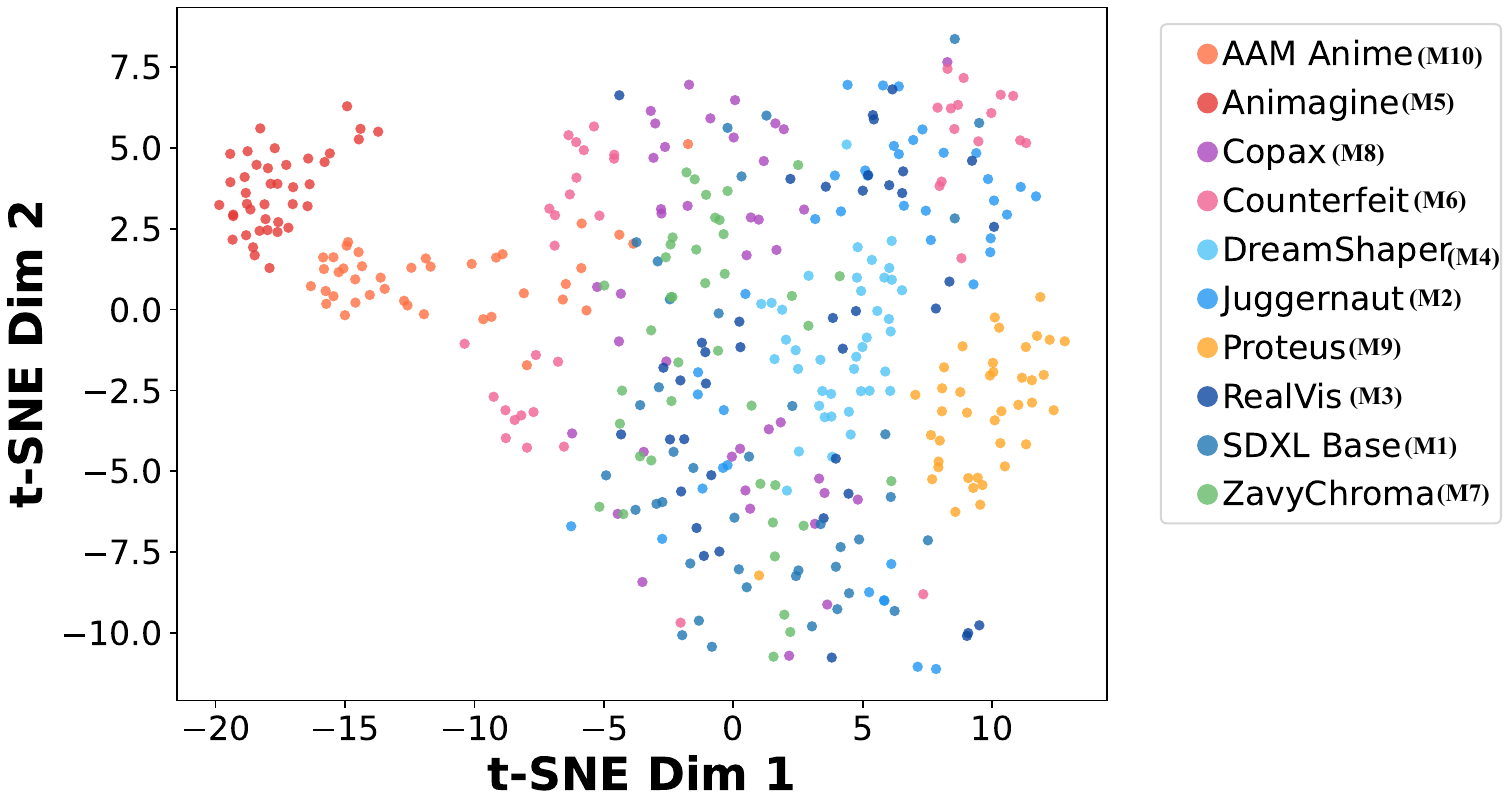}
    \caption{t-SNE projection of frozen CLIP ViT-L/14 features of input images generated using a suitable shared prompt across all 10 SDXL fine-tunes.}  
    \label{fig:clip_tsne}
\end{figure}

\section{Style Capture using Pretrained CLIP} \label{sec:clip}
In Section 4.1 of the \textit{main paper}, we state that frozen CLIP~\cite{radford2021learning} would be able to capture visual style representative features from the input image.
To empirically verify this property on our ISDB dataset, we visualize the t-SNE projection of the frozen CLIP features extracted from all the 10 SDXL fine-tunes using the same single prompt suitable for all in Fig.~\ref{fig:clip_tsne}. 
So, for an input image $\mathcal{I}$, we take a $518 \times 518$ crop and resize it to $224 \times 224$ for CLIP, which produces a style representation feature $\mathbf{x}_{\text{style}} \in \mathbb{R}^{d_{\text{clip}}}$ where $d_{\text{clip}} = 1024$. The t-SNE plot in the figure shows the projection of $\mathbf{x}_{\text{style}}$ for the input image as a point. As can be seen, the anime-styled images generated from fine-tuned SDXL anime image generators Animagine (M5), Counterfeit (M6) and AAM Anime (M10) are reasonably separated from the images generated by stylized (M7, M8, M9) and photorealistic (M1, M2, M3, M4) fine-tuned generators derived from SDXL (see Table~1 in the \textit{main paper} for the style categories of the SDXL fine tunes considered). The stylized and photorealitic generated images are somewhat separated with former mostly occupying the bottom \& middle, and the latter occupying the middle \& top-right parts of the plot. Therefore, although we visualize the 1024-d vectors in an optimized 2-d view, we still find some impression of separation between them in terms of image style.


\section{Intra-architecture Attribution Results}
\label{sec:ExistingAttribution}

\begin{table*}[!t]
\centering
\caption{Mean Mixture of Expert (MoE) gating weights (\%) per model, grouped \& averaged by style type.}
\label{tab:expert_gating}
\begin{tabular}{|l|ccccc|cccc|cccc|}
\hline
\multirow{2}{*}{ Experts}& \multicolumn{5}{c|}{Photorealistic} & \multicolumn{4}{c|}{Stylized} & \multicolumn{4}{c|}{Anime}\\ \cline{2-14}
& M4 & M2 & M1 & M3 & avg. & M8 & M7 & M9 & avg. & M6 & M10 & M5 & avg.  \\
\hline
Expert 1 & 35.5 & 36.7 & 37.1 & 37.6 & 36.7 & 37.2 & 36.6 & 34.9 & 36.2 & 33.5 & 28.1 & 15.8 & 25.8\\
Expert 2 & 24.9 & 24.8 & 23.0 & 25.2 & 24.5 & 26.3 & 26.8 & 27.2 & 26.8 & 27.2 & 29.2 & 27.4 & 27.9\\
Expert 3 & 23.7 & 21.6 & 23.6 & 21.2 & 22.5 & 18.8 & 19.5 & 19.6 & 19.3 & 17.8 & 17.4 & 13.3 & 16.2\\
Expert 4 & 15.9 & 16.9 & 16.2 & 16.1 & 16.3 & 17.7 & 17.1 & 18.3 & 17.7 & 21.6 & 25.3 & 43.6 & 30.2\\
\hline
\end{tabular}
\end{table*}

\begin{table}[t]
\centering
\caption{Cross-Dataset Performance of Our \textit{IntrAA-SD} baseline}
\label{tab:cross}
\begin{tabular}{c|c|c|c}
\hline
 \multirow{2}{*}{\textbf{Training Set}}& \multicolumn{3}{c}{\textbf{ Test Set Accuracy\%}} \\ \cline{2-4}
 & \textbf{ISDB-ID} & \textbf{ImageNet} & \textbf{COCO} \\
\hline
ImageNet & 80.59 & 98.35 & 96.47\\
COCO & 76.16 & 90.31 & 98.91\\
ISDB-ID (Ours) & 99.03 & 84.58 & 85.95\\
\hline
\end{tabular}
\end{table}

\begin{table*}[t]
\centering
\caption{Confusion Matrix related to performance of our \textit{IntrAA-Diff} on ISDB-ODD set. Photorealistic: \textbf{M1-4}, Stylized: \textbf{M7-9}, Anime: \textbf{M5-6, M10}.}
\label{tab:supp_cm}
\begin{tabular}{l | cccccccccc}
\hline
\textbf{True $\downarrow$ / Pred $\rightarrow$} & \textbf{M1} & \textbf{M2} & \textbf{M3} & \textbf{M4} & \textbf{M5} & \textbf{M6} & \textbf{M7} & \textbf{M8} & \textbf{M9} & \textbf{M10} \\
\hline
\textbf{M1 (SDXL Base)}      & \textbf{923} & 3 & 2 & 0 & 1 & 5 & 11 & 46 & 5 & 4 \\
\textbf{M2 (Juggernaut)}     & 22 & \textbf{708} & 149 & 8 & 0 & 14 & 49 & 28 & 16 & 6 \\
\textbf{M3 (RealVis)}        & 16 & 87 & \textbf{816} & 4 & 0 & 12 & 35 & 18 & 9 & 3 \\
\textbf{M4 (DreamShaper)}    & 0 & 5 & 0 & \textbf{985} & 0 & 0 & 4 & 0 & 1 & 5 \\
\textbf{M5 (Animagine)}      & 4 & 0 & 0 & 0 & \textbf{984} & 1 & 3 & 0 & 0 & 8 \\
\textbf{M6 (Counterfeit)}    & 1 & 3 & 0 & 1 & 2 & \textbf{978} & 5 & 0 & 3 & 7 \\
\textbf{M7 (ZavyChroma)}     & 57 & 45 & 43 & 4 & 1 & 15 & \textbf{742} & 50 & 25 & 18 \\
\textbf{M8 (Copax)}          & 51 & 3 & 4 & 2 & 0 & 6 & 20 & \textbf{908} & 4 & 2 \\
\textbf{M9 (Proteus)}        & 8 & 16 & 12 & 3 & 5 & 1 & 7 & 8 & \textbf{940} & 0 \\
\textbf{M10 (AAM Anime)}     & 6 & 1 & 0 & 5 & 8 & 7 & 24 & 10 & 0 & \textbf{939} \\
\hline
\end{tabular}
\end{table*}


Table~\ref{tab:main_results} lists the intra-architecture source attribution results obtained for our \textit{IntrAA-SD} baseline framework on our ISDB dataset, which is also reported in Table 4 of the \textit{main paper}. However, in Table~\ref{tab:main_results} here, the performance of five other well-known recent methods, which are designed to perform architecture-level source attribution including diffusion-based generators, are reported. 
As shown in the table, when existing architectural-level source attribution models are trained on our training set, the best average accuracies achieved on our test set are 92.21\% and 74.84\% for ID and OOD cases, respectively.  Their performance remains substantially below  our intra-architecture source attribution baseline, which  is specifically designed for intra-architecture attribution.  This suggests that the development of dedicated intra-architecture attribution techniques is necessary continuing from our baseline framework, particularly to improve generalization on the OOD test set of the ISDB dataset. 


\section{Additional Experiments} \label{sec:experiments}
\subsection{Mixture of Experts (MoE) Gating Weights}\label{sec:routing}
Table~\ref{tab:expert_gating} reports the mean MoE gating weights for each expert per fine-tuned model, and the average over the models for a particular style-type is also provided. As can be seen, the Experts 1-4 respectively contribute on an average around 37\%, 25\%, 23\%, and 16\% for the photorealistic models, whereas they in that order contribute on an average 36\%, 27\%, 19\%, \& 18\%, and 26\%, 28\%, 16\%, \& 30\% for the stylized and Anime models, respectively. Therefore, the distribution of the average contributions by the four experts change with the style-type of the models. This signifies that differently-styled generated images might have structurally different source model signatures as stated in Section~4.3 of the \textit{main paper}, which can be captured by learning different distribution of weighting multiple experts for computing model trace features.


\subsection{Cross-dataset Intra-Architecture Attribution Performance}
Datasets like ImageNet~\cite{deng2009imagenet} and COCO~\cite{lin2014microsoft}, which contain captions along with images, have primarily been used for object /image recognition purpose. Here, we study the appropriateness of using such datasets to train intra-architecture source attribution models instead of using specially curated image \& prompt set like the ID set in ISDB dataset (ISDB-ID), and also perform cross-dataset performance analys of our \textit{IntrAA-SD} baseline approach for intra-architecture attribution.

So, we first prepare two datasets from the $10$ SDXL fine-tuned models using the image-caption pairs from ImageNet and COCO datasets for training two variants of our \textit{IntrAA-SD} baseline approach. The first dataset includes 12,000 images per fine-tuned model generated using ImageNet captions as prompts, and the second has 12,000 images per fine-tuned model generated using the COCO captions. We split the each set into 80/20 train and test partitions using stratified sampling with a fixed random state, ensuring no prompt overlap between splits. We test our ISDB-ID trained \textit{IntrAA-SD} baseline on these datasets. We also test our \textit{IntrAA-SD} baseline variants, which are trained from scratch on the two training sets formed from ImageNet and COCO datasets, on our ISDB-ID test set.

\begin{figure*}[!t]
  \centering
  \begin{tabular}{@{}c@{\hskip 2pt}c@{\hskip 2pt}c@{\hskip 2pt}c@{}}
    \includegraphics[width=0.22\textwidth]{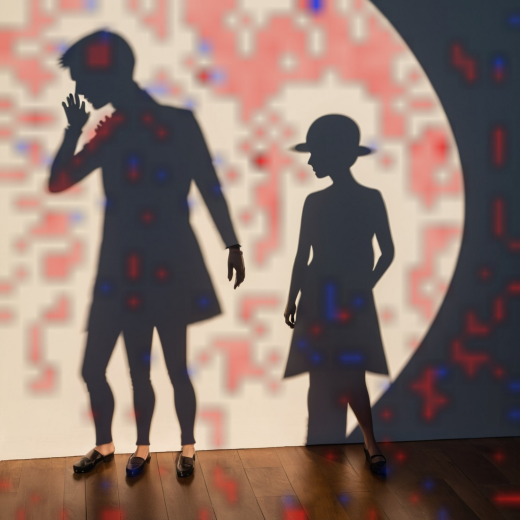} &
    \includegraphics[width=0.22\textwidth]{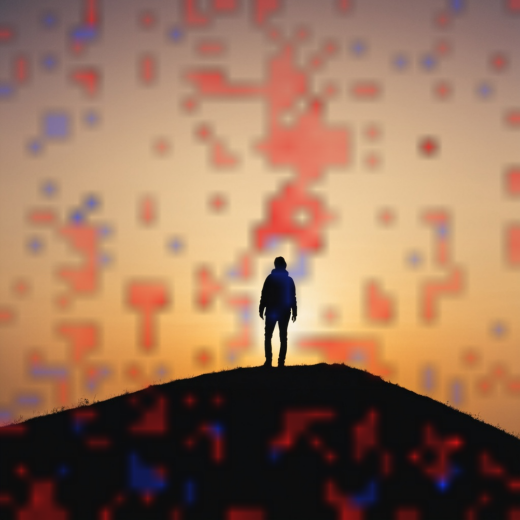} &
    \includegraphics[width=0.22\textwidth]{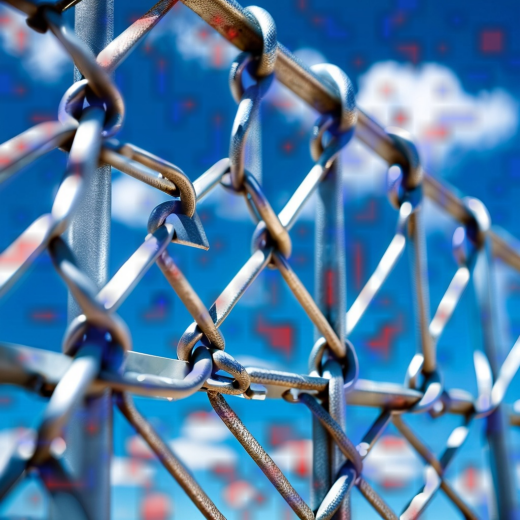} &
    \includegraphics[width=0.22\textwidth]{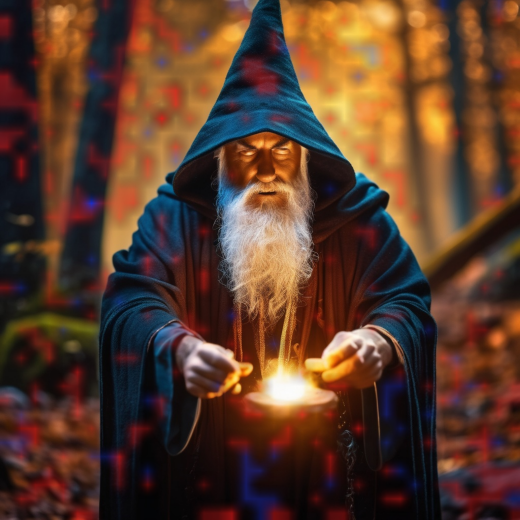} \\
  \end{tabular}
  \caption{Significant changes (Red: increase, Blue: decrease) at patch-level feature norm after LoRA adaptation of DINOv2 for images from four different fine-tuned models.}
  \label{fig:feature_norms}
\end{figure*}
The results of the cross-dataset experiment is shown in Table~\ref{tab:cross}. As can be seen, when the training and testing of our \textit{IntrAA-SD} baseline are done on the same dataset, the performance achieved is quite high, irrespective of the datasets. However, the performance drops to different extents in the different cases when the the training and testing are done on different datasets. If the training and testing are done on ImageNet and COCO, one or the other, the performance achieved is still above $90\%$, which signifies the similarity of two datasets derived from them. If the training is done on ImageNet or COCO derived datasets, and the testing on our ISDB-ID test set, the performance barely touches $80\%$. Whereas when the training is done on our ISDB-ID train set, and the testing on ImageNet or COCO, the performance manages to reach around $85\%$. 

These observations suggests that the cross-dataset performance of our \textit{IntrAA-SD} baseline is just satisfactory leaving scope for future improvement. Further, another important finding is that training on our ISDB-ID train set imbibes our baseline approach with slightly better generalization capability achieving around $5\%$ accuracy boost. This highlights the importance of such specially curated datasets for intra-architecture source attribution.


\subsection{Confusion Matrix of \textit{IntrAA-SD} baseline on OOD set in ISDB dataset}
From Table~4 in the \textit{main paper}, we see that our \textit{IntrAA-SD} baseline trained on our ISDB-ID training set achieves 88.55\% attribution accuracy on the OOD test set of the ISDB dataset (ISDB-OOD).
We further examine here the related confusion matrix in Table~\ref{tab:supp_cm}. A salient observation from the matrix is that images generated from a majority of the photorealistic and stylized image generators are prone to be misclassified as from the SDXL-Base source. This is understandable as SDXL-Base is categorized under photorealistic generator producing images with styles closer to other photorealistic and stylized category fine-tunes rather than the anime category ones. In fact, images from the fine-tune ZavyChroma (M7) in the stylized category has got misclassified as obtained from three of the four photorealistic fine-tunes. Another key observation is that slightly higher (around $8-15\%$) misclassifications has occurred just between photorealistic fine-tunes Juggernaut (M2) and RealVis (M3), as they produce images with very similar visual styles. The above findings point to paths for designing improved intra-architecture attribution methods in future.


\subsection{Effect of LoRA on Feature Representations}\label{sec:loraeffect}
As mentioned in Section~4.2 of the \textit{main paper}, our \textit{IntAA-SD} baseline model contains a LoRA-adapted DinoV2 to extract fine-tuned model's trace features from the input image. To visualize the DINOv2's internal representation modification by LoRA adaptation, we compute the $L_2$ norm of each patch token from the final transformer layer for both frozen DINOv2 and LoRA-adapted DINOv2. For each of the $37 \times 37$ spatial patches ($\mathbf{h}_i$ vector), we measure the relative change $\Delta_i = (\lVert \mathbf{h}_i^{\text{\tiny LoRA+DINO}} \rVert_2 - \lVert \mathbf{h}_i^{\text{\tiny frozen\_DINO}} \rVert_2) / \lVert \mathbf{h}_i^{\text{\tiny frozen\_DINO}} \rVert_2$ and mark only the patches in the top 25\textsuperscript{th} percentile of $|\Delta_i|$ for visualization in  Fig.~\ref{fig:feature_norms}. Among those, positive $\Delta_i$ patches are shown in red, which can be seen is mostly concentrated on smooth and less-textured regions. The negative $\Delta_i$ patches are shown in blue, which we see appear in all regions including salient regions like faces, knots, and edges \&corners. Such observations are found to be consistent across images from the different fine-tuned generators. Therefore, LoRA seems to increase (red patches) the magnitude of the features in the smooth, low-texture patches, indicating more likelihood of discernible model signature presence in these regions. Analogously, it also suppresses (blue patches) the magnitude of features at a variety of regions which include salient ones, where model signature might be well-camouflaged in general.



{
    \small
    \bibliographystyle{ieeenat_fullname}
    \bibliography{main}
}

%% file: sec/0_abstract.tex
\begin{abstract}
Designing systems that attribute a generated image to its source model has lately become a major area of AI research. Most existing datasets and systems related to image source attribution are targeted towards the detection of architectural signature of the source models. Distinct from this, focusing on stable diffusion, we introduce a novel task, \textbf{IntrAA-SD}, \textbf{Intr}a-\textbf{A}rchitecture source \textbf{A}ttribution for \textbf{S}table \textbf{D}iffusion, where the source models to be differentiated share the same architectural backbone and differ only due to fine-tuning. The proposed task aligns with the emergence of fine-tuned generators having the same architecture, like Juggernaut XL and RealVis XL which are fine-tuned versions of Stable Diffusion XL (SDXL). We provide a first-of-its-kind $140$K-image benchmark dataset containing images generated from different fine-tunes of SDXL, along with in-domain and out-of-distribution prompts. We also provide a baseline framework for IntrAA-SD, which leverages a style-conditioned Mixture-of-Expert network. It uses a frozen CLIP backbone to capture the image style and gate the MoE, which processes fine-tunes' trace features from a high-resolution LoRA-adapted DINOv2 network. The fine-tuned models used, the prompting strategies considered, and the formation details employed in the novel dataset are elaborated. Our baseline intra-architecture source attribution method is evaluated in terms of attribution accuracy on our dataset, and ablation studies are also performed. Code and dataset is available \href{https://github.com/mallikagarg/IntrAA-SD}{\underline{here}}.
\end{abstract}

%% file: sec/1_intro.tex
\vspace{-4mm}
\section{Introduction}
\label{sec:intro}
Recent developments in generative AI has led to the availability of many open-weight text-to-image foundational models and their numerous fine-tuned versions. Therefore, attribution of image generations, not only to a specific architectural model but also to its particular fine-tuned version, is required for proper crediting and to identify misuse~\cite{qu2023unsafe}.
Although source attribution systems and related datasets are available that aim to link a generated image to an architectural model, the advent of fine-tuned generators has made reliable source attribution to specific fine-tunes difficult under the existing frameworks~\cite{deepfake2025}.


Branching off from the task of detecting whether an image is fake or not~\cite{wang2020}, the field of source attribution emerged for identifying the specific models that generated the given images~\cite{yang2022deepfake} with the goal to facilitate intellectual property protection~\cite{yu2021artificial}.
Recently, diffusion-based generators such as DALL-E~\cite{betker2023improving}, Midjourney~\cite{midjourney_github}, Imagen~\cite{saharia2022photorealistic} and the Stable Diffusion~\cite{rombach2022high} family have made high-quality synthetic image generation widely accessible, and hence, source attribution for diffusion-based generative models have since been investigated. Such architecture-level source attribution considers the classification of images from generator models that have significantly different architectures and generation pipelines~\cite{yu2019attributing, frank2020leveraging, yang2022deepfake, sha2023fake, corvi2023, guarnera2024, Fei_2026_CVPR, yan2025orthogonal}, like DALL-E 3 and Midjourney v6. 
However, models like Stable Diffusion XL (SDXL)~\cite{sdxl} 
have a multitude of fine-tuned versions, which have been effectively used to generate images of various kinds, ranging from anime art to photorealistic portraits. The existing frameworks of designing systems for source attribution by detecting architectural signatures with the help of training on datasets of images created using different diffusion-based generators are neither intended nor suited for attributing the generation to such specific fine-tunes.

\begin{figure*}[!t]
\centering
\setlength{\tabcolsep}{0.5pt}
\renewcommand{\arraystretch}{0.3}
\scriptsize
\newcommand{\thumbw}{0.093\textwidth}
\begin{tabular}{c cccccccccc}
& \rotatebox{0}{\;\;M1}
& \rotatebox{0}{\;\;M2}
& \rotatebox{0}{\;\;M3}
& \rotatebox{0}{\;\;M4}
& \rotatebox{0}{\;\;M5}
& \rotatebox{0}{\;\;M6}
& \rotatebox{0}{\;\;M7}
& \rotatebox{0}{\;\;M8}
& \rotatebox{0}{\;\;M9}
& \rotatebox{0}{\;\;M10} \\[4pt]
P1 &
\includegraphics[width=\thumbw]{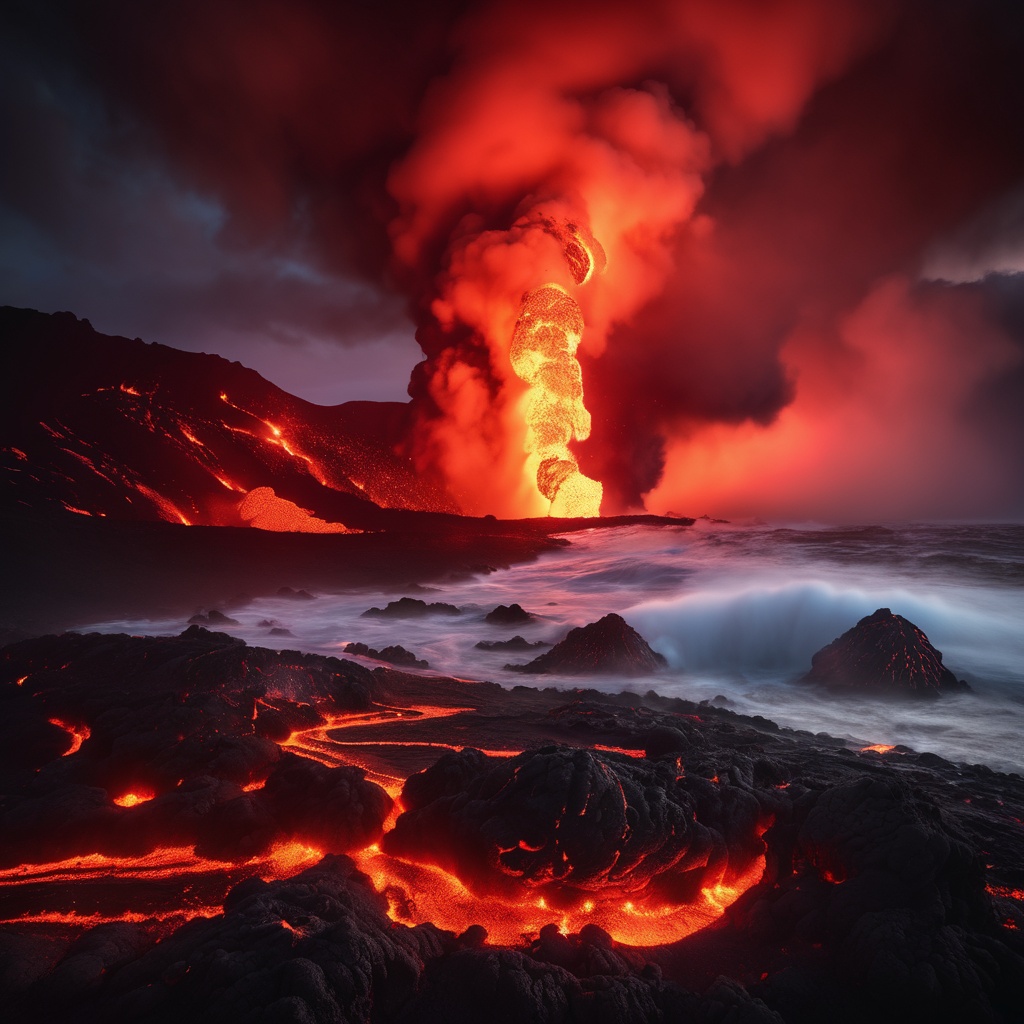} &
\includegraphics[width=\thumbw]{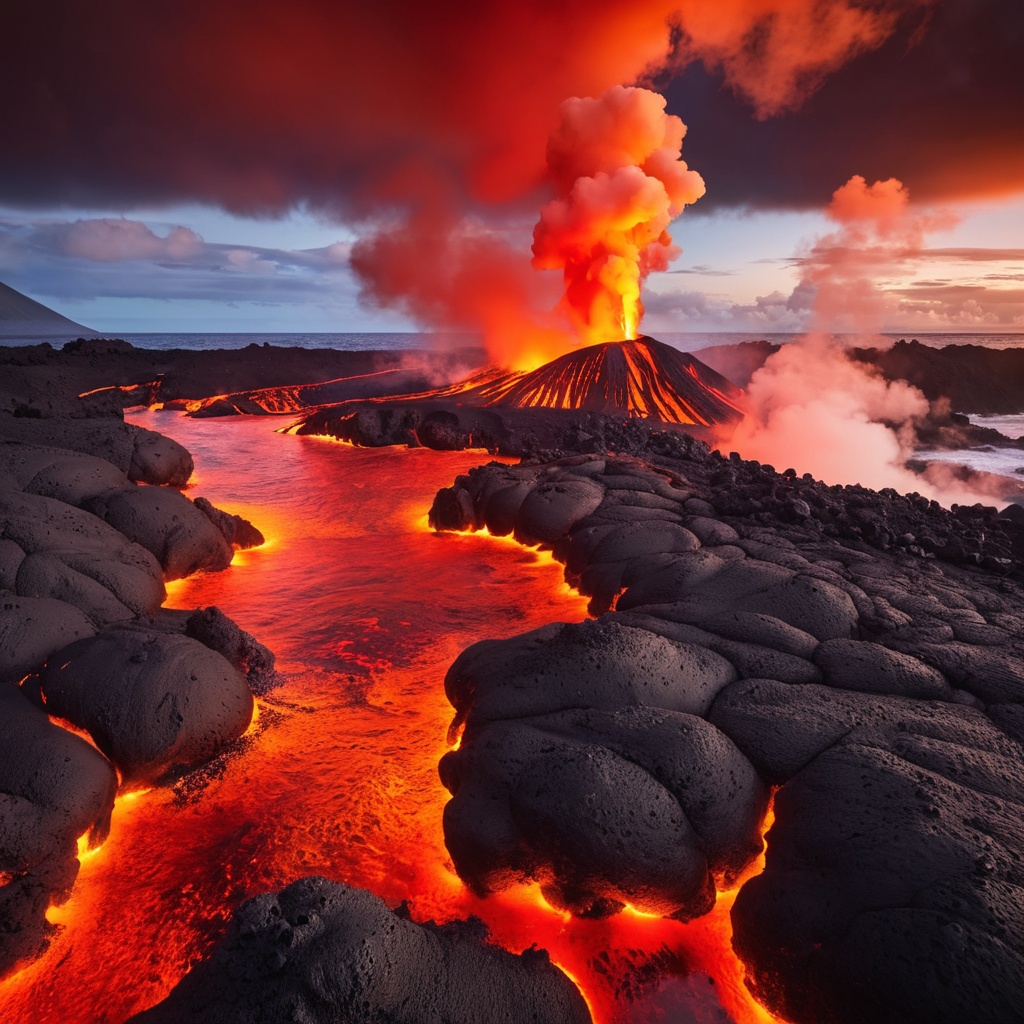} &
\includegraphics[width=\thumbw]{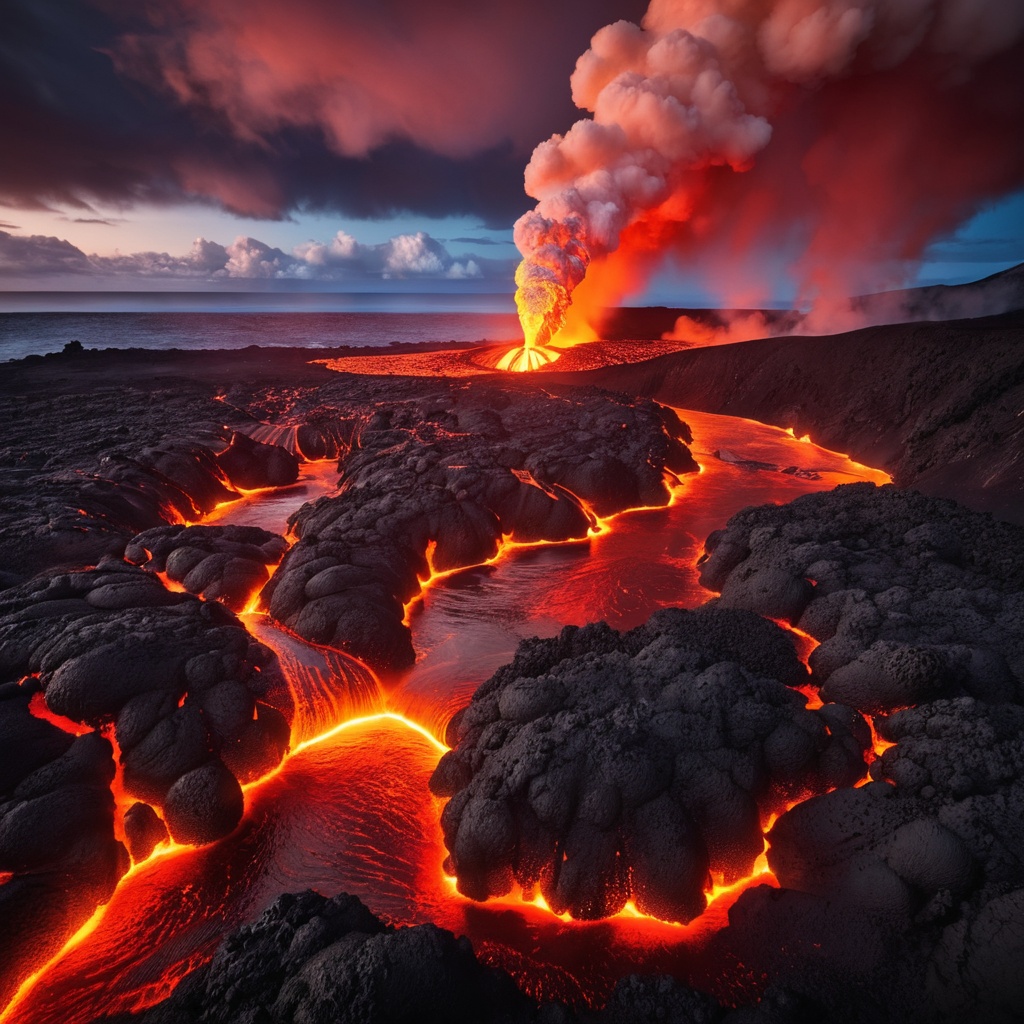} &
\includegraphics[width=\thumbw]{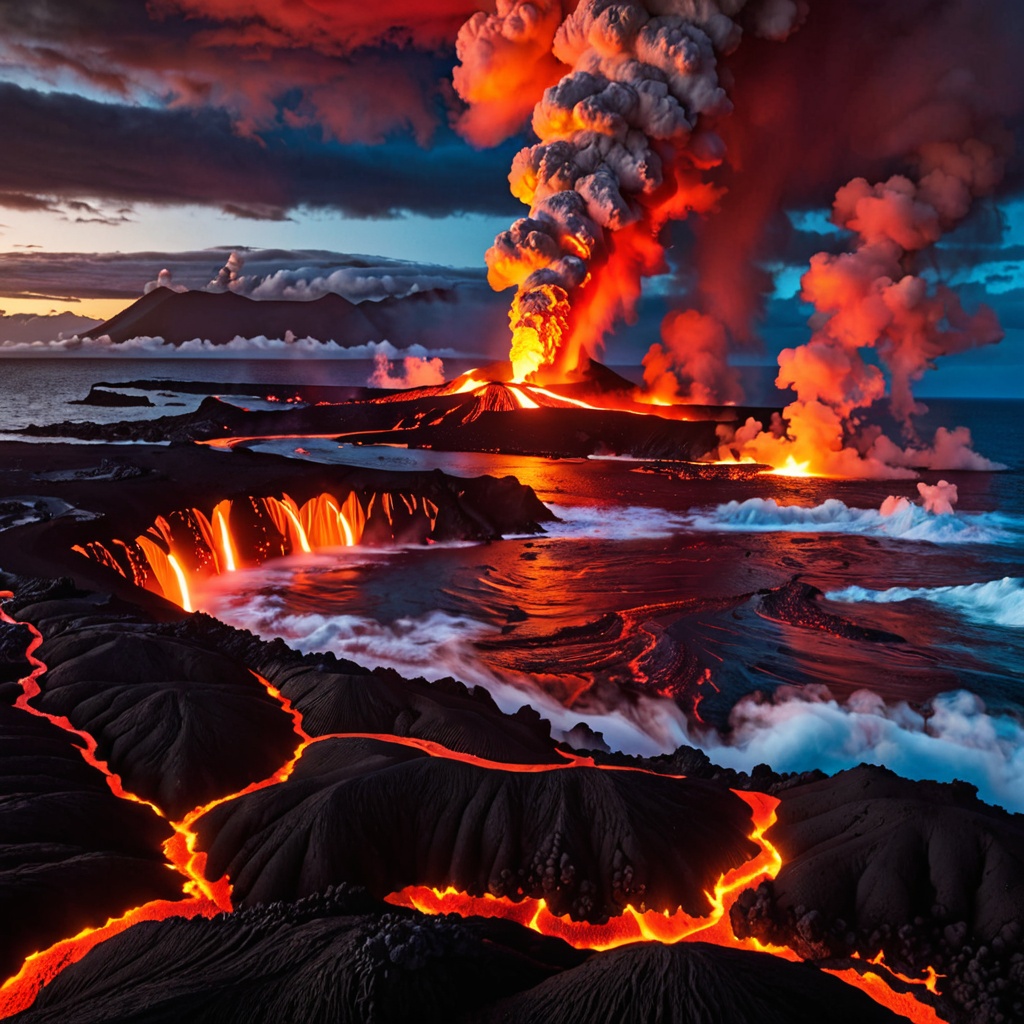} &
\includegraphics[width=\thumbw]{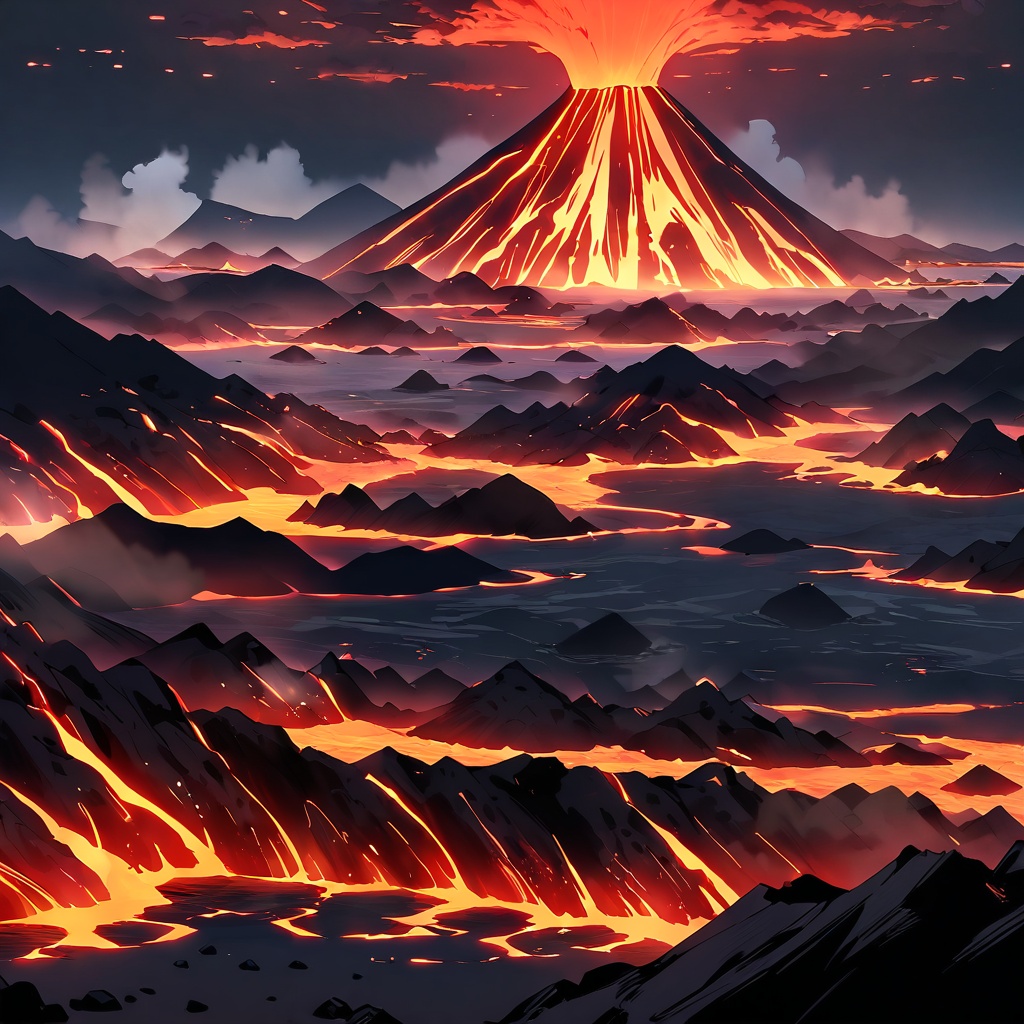} &
\includegraphics[width=\thumbw]{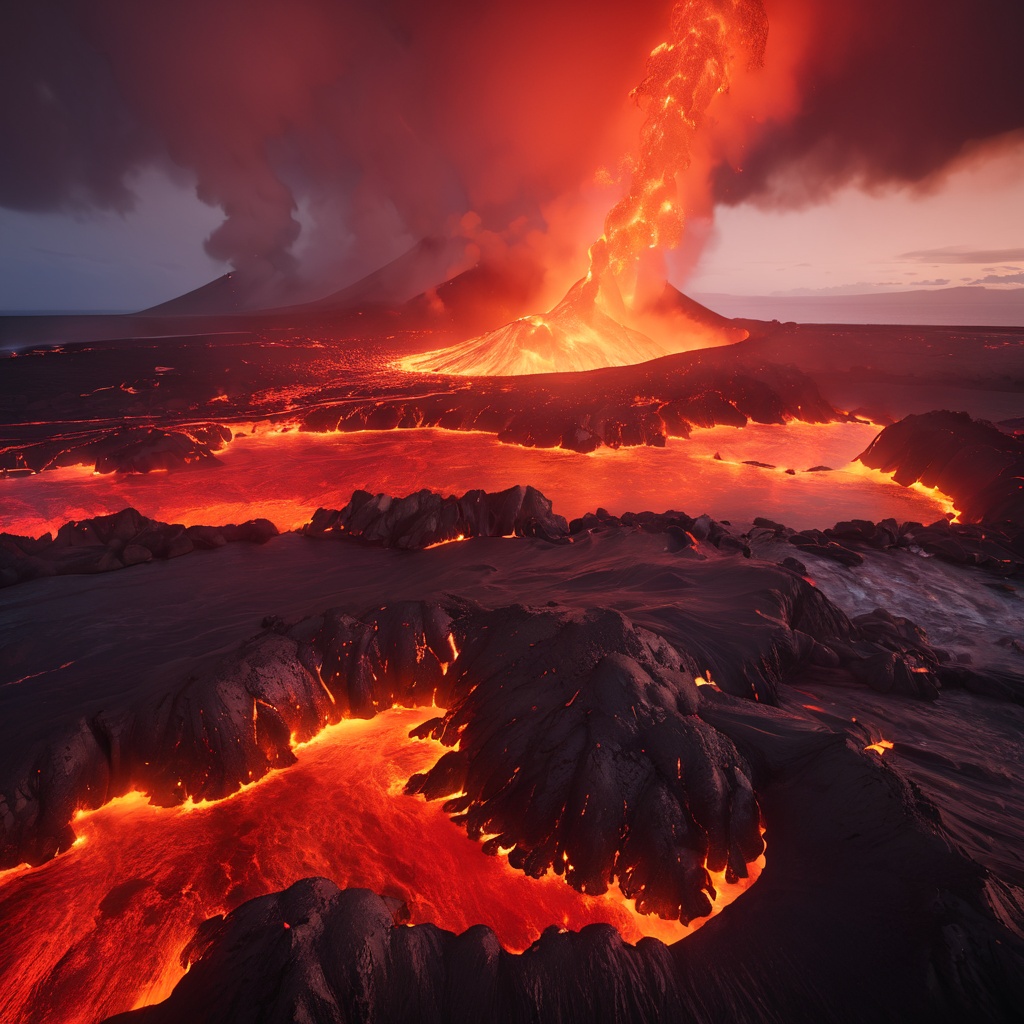} &
\includegraphics[width=\thumbw]{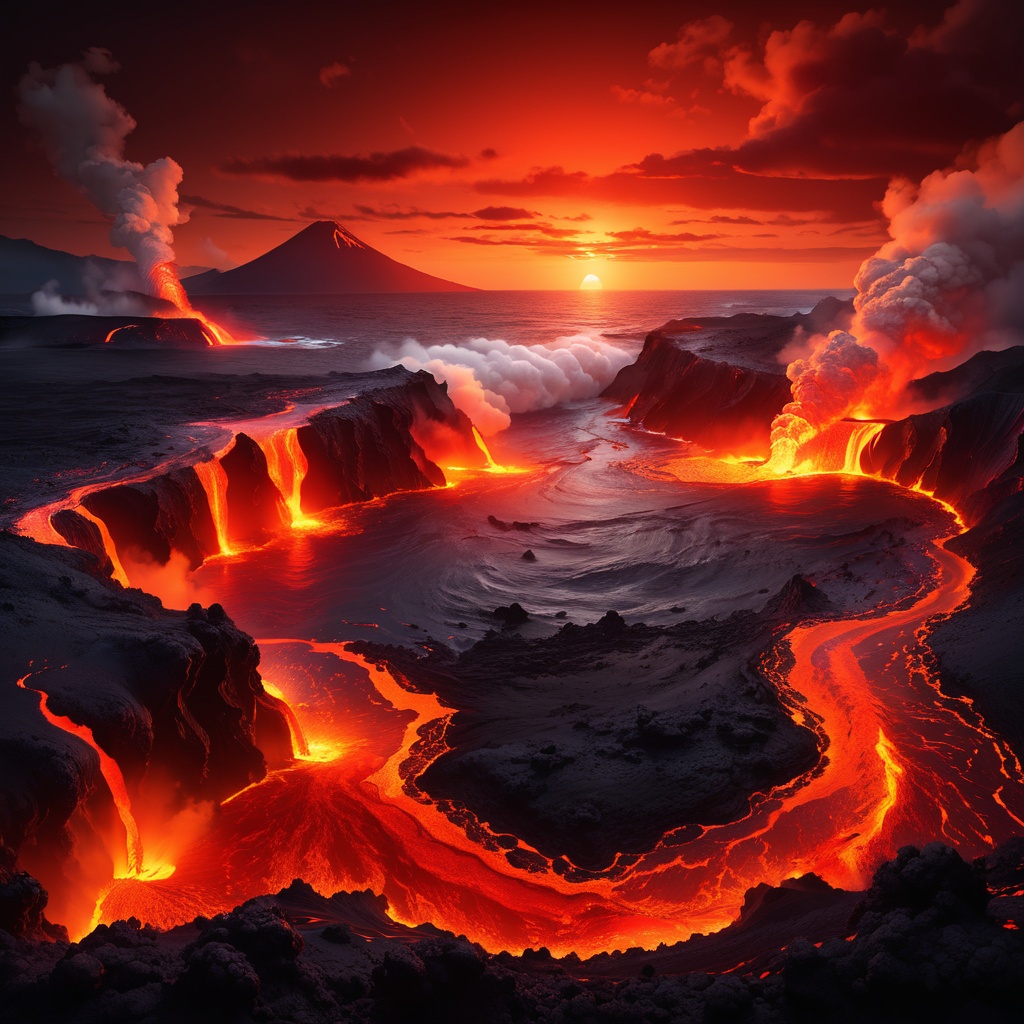} &
\includegraphics[width=\thumbw]{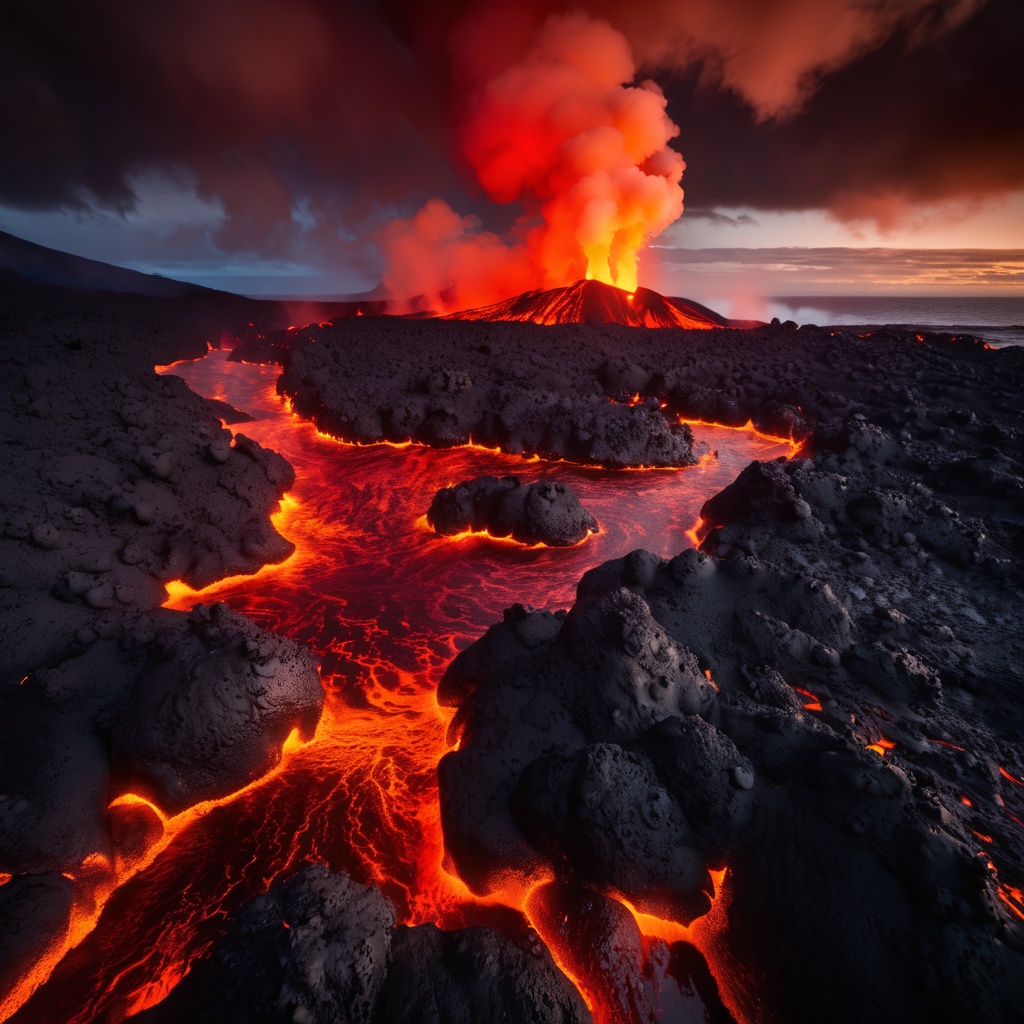} &
\includegraphics[width=\thumbw]{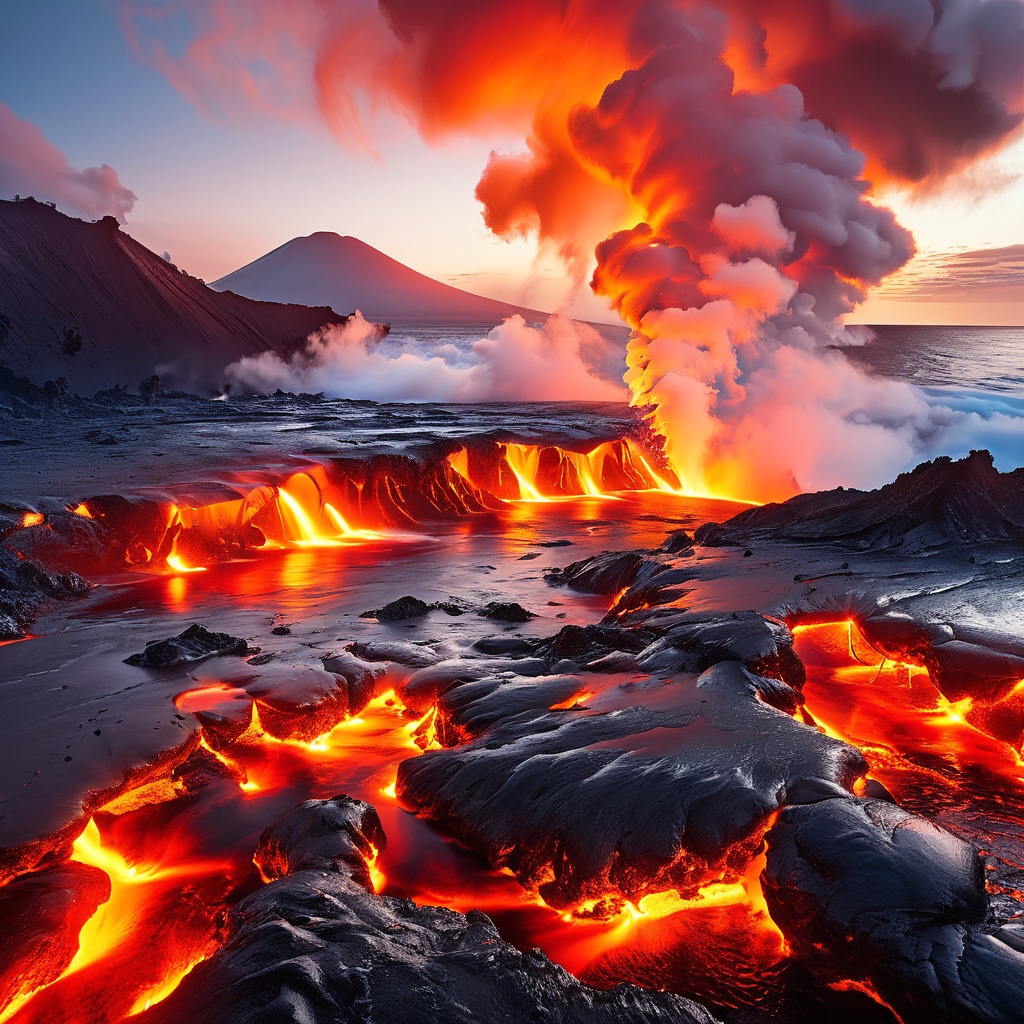} &
\includegraphics[width=\thumbw]{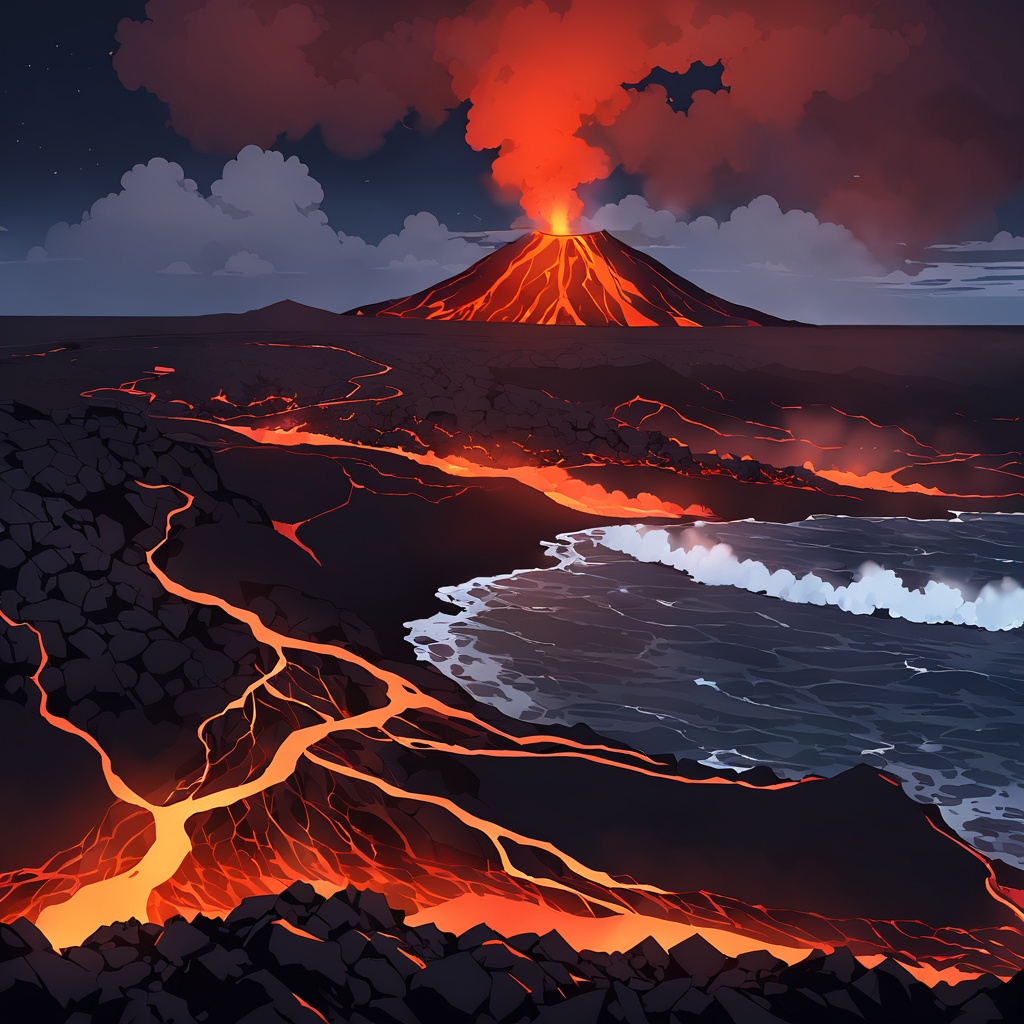} \\
P2 &
\includegraphics[width=\thumbw]{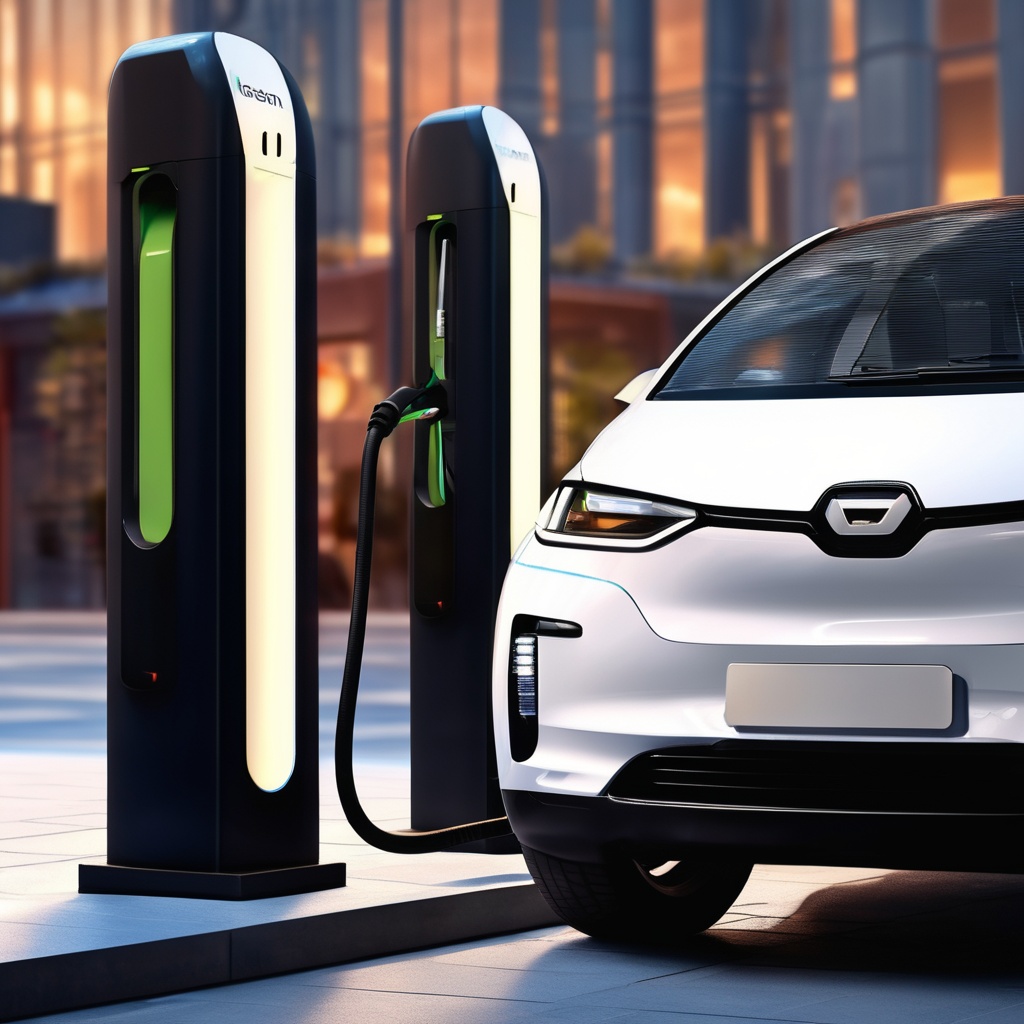} &
\includegraphics[width=\thumbw]{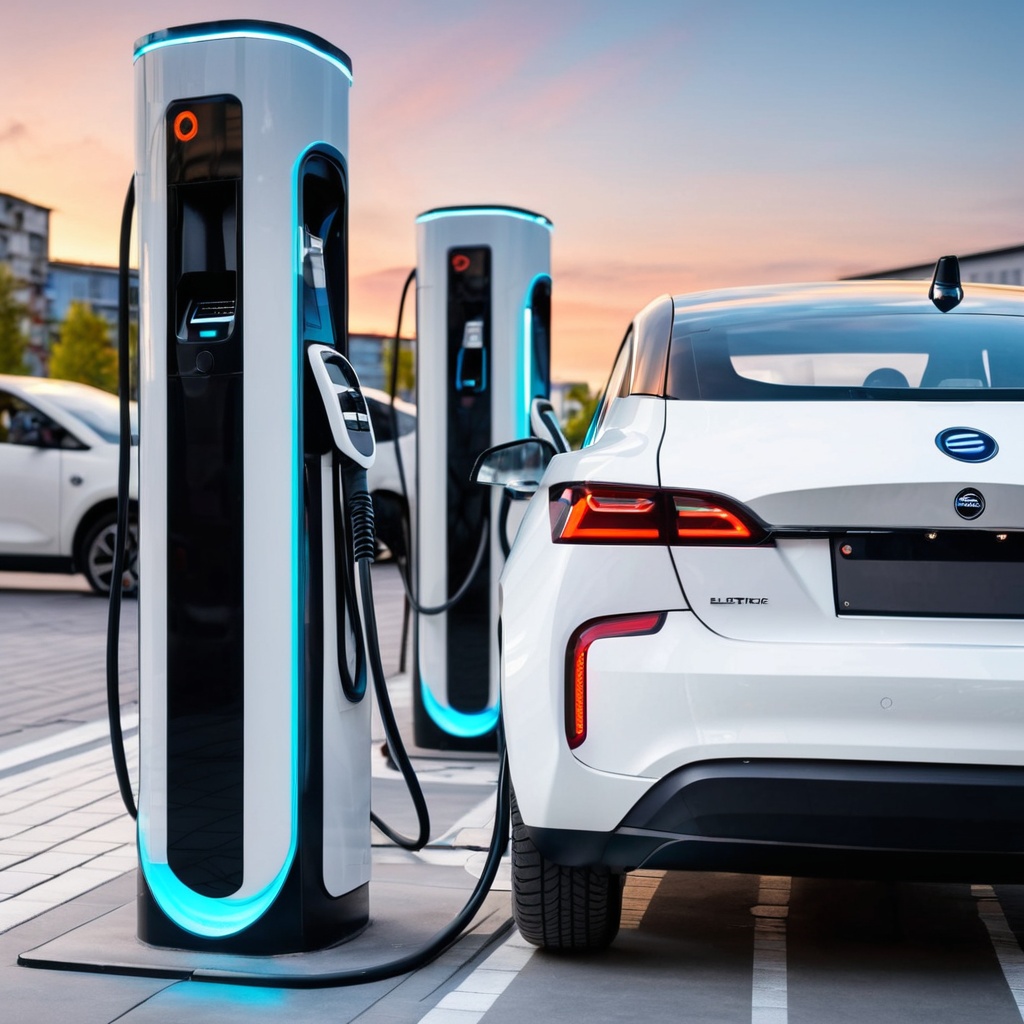} &
\includegraphics[width=\thumbw]{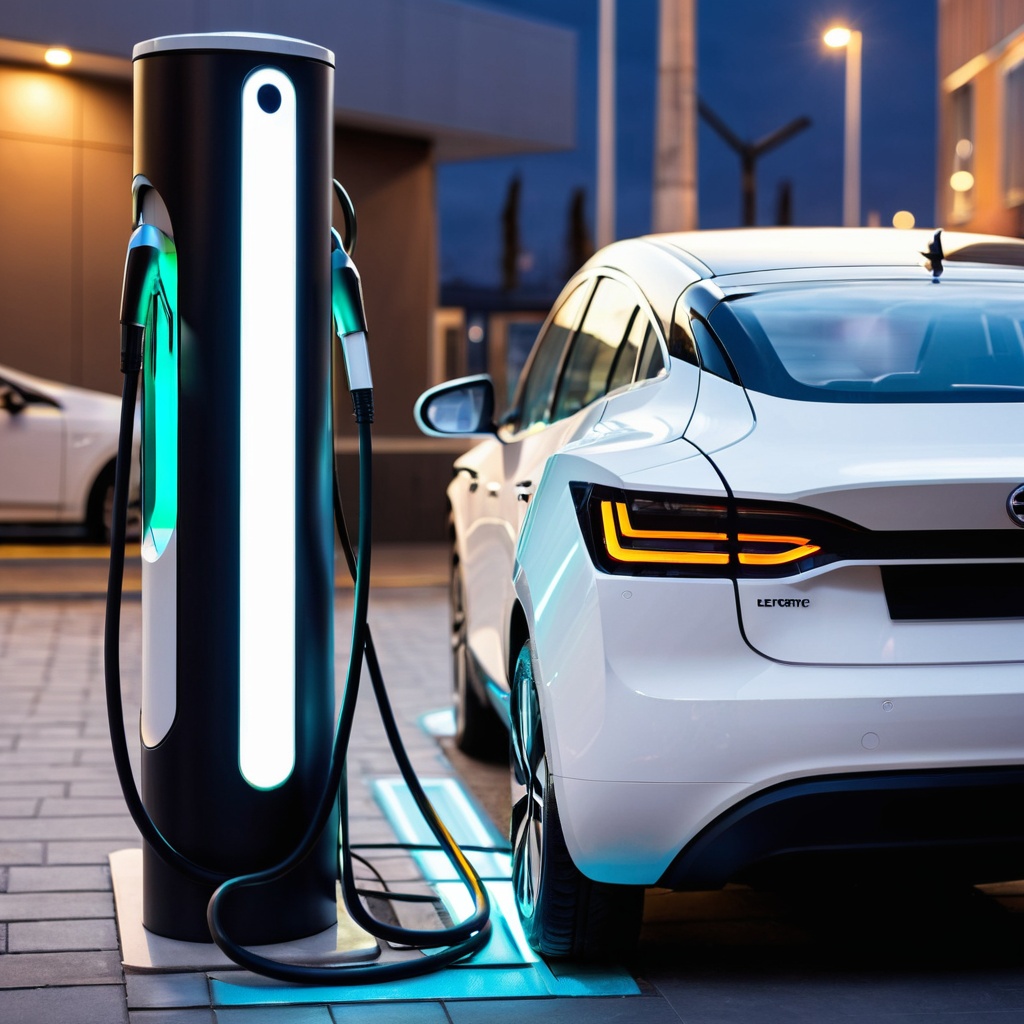} &
\includegraphics[width=\thumbw]{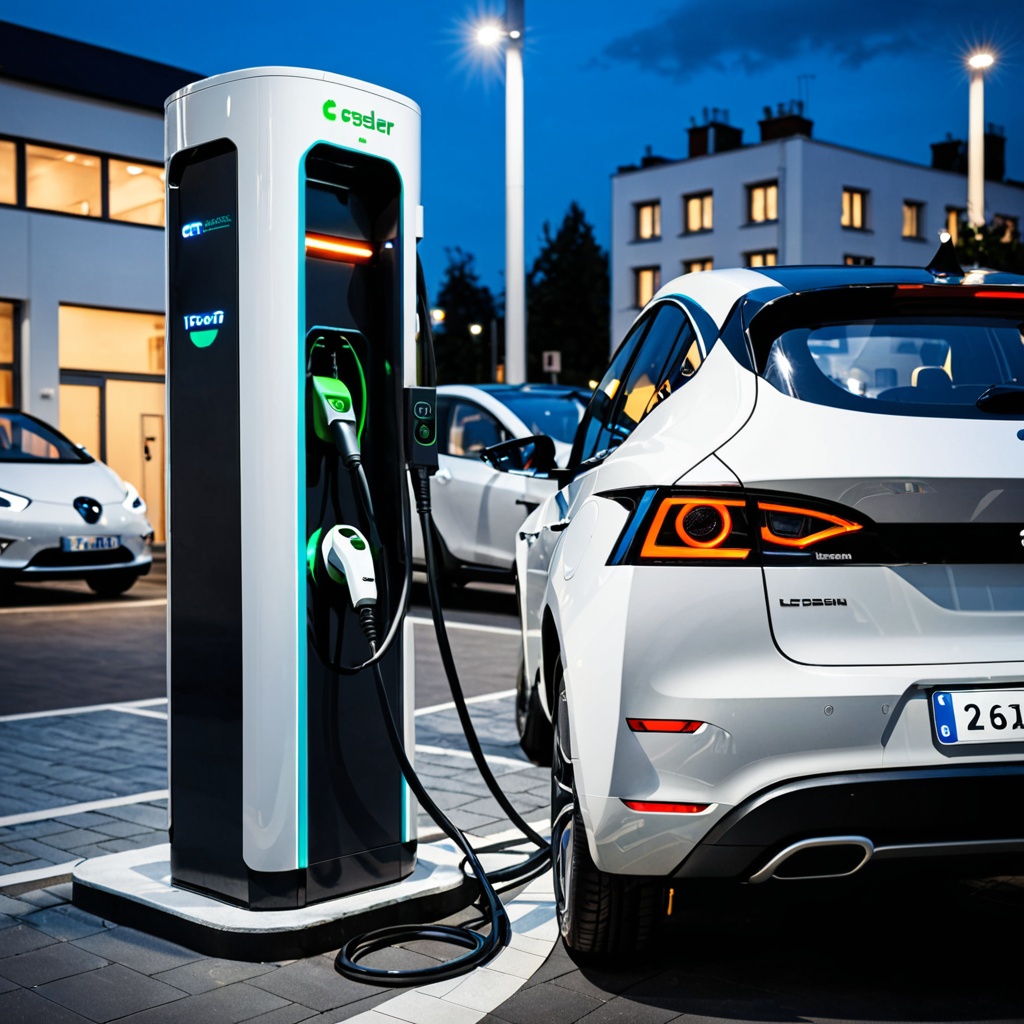} &
\includegraphics[width=\thumbw]{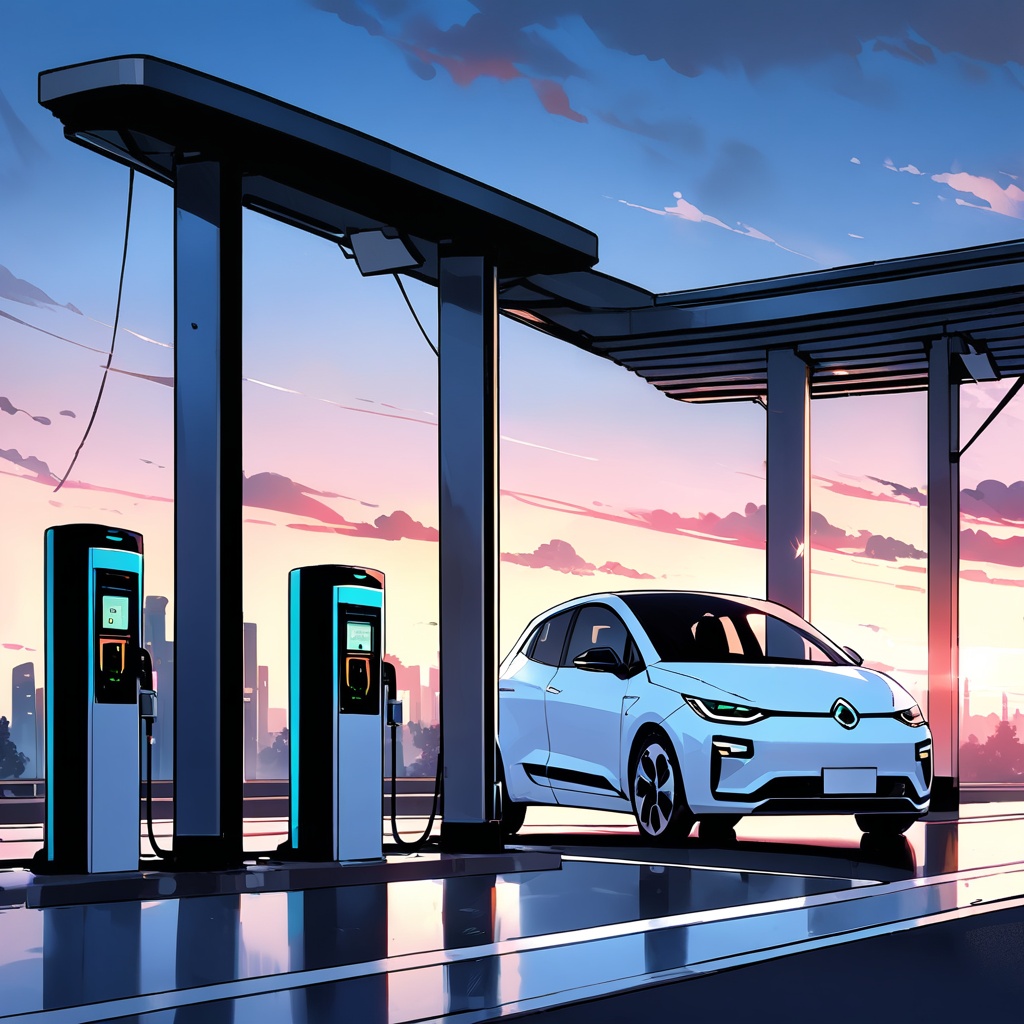} &
\includegraphics[width=\thumbw]{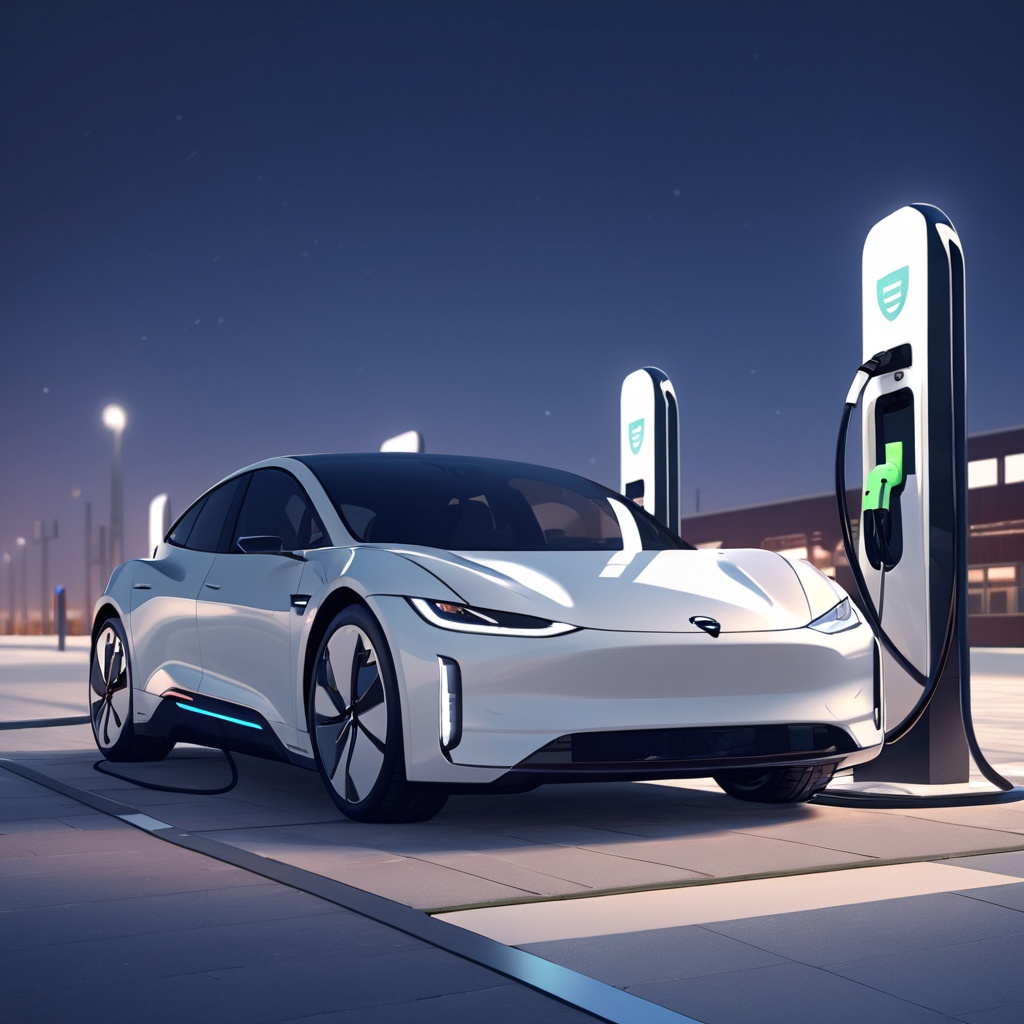} &
\includegraphics[width=\thumbw]{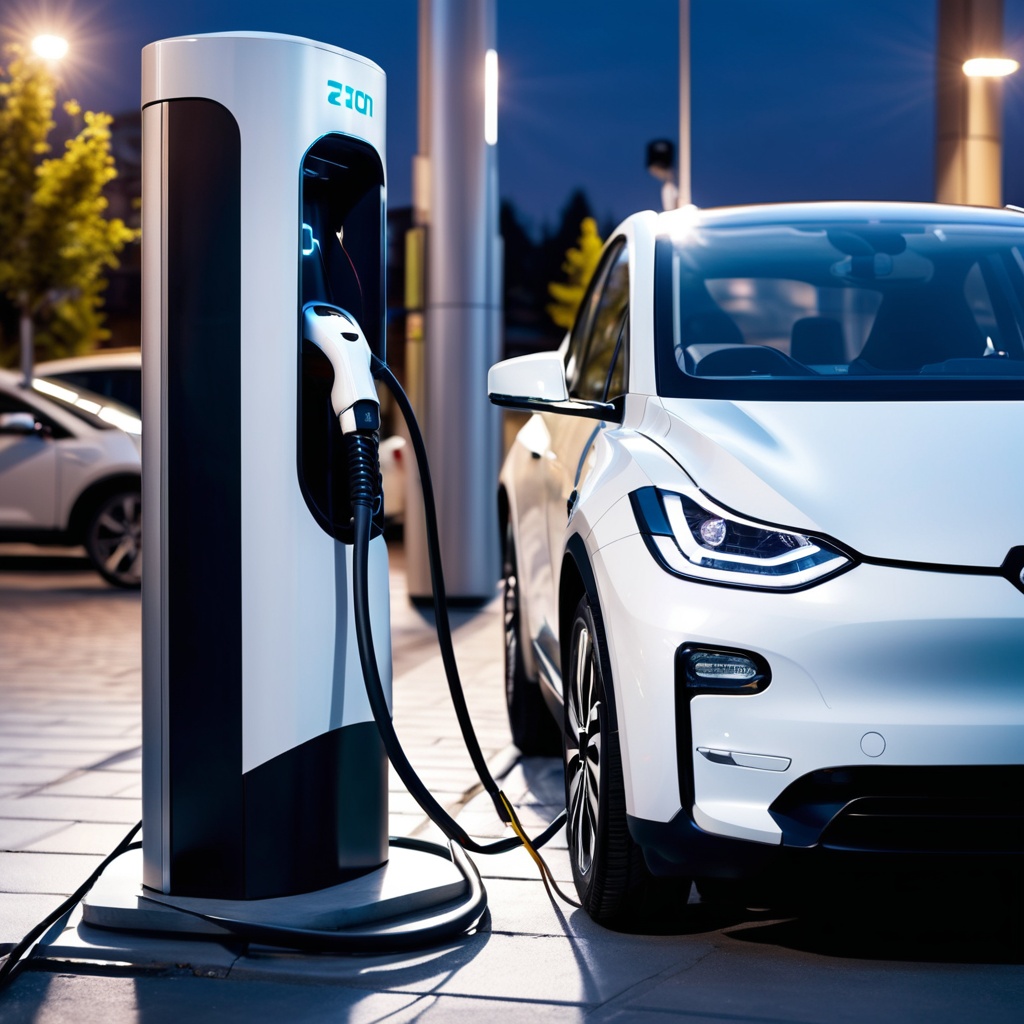} &
\includegraphics[width=\thumbw]{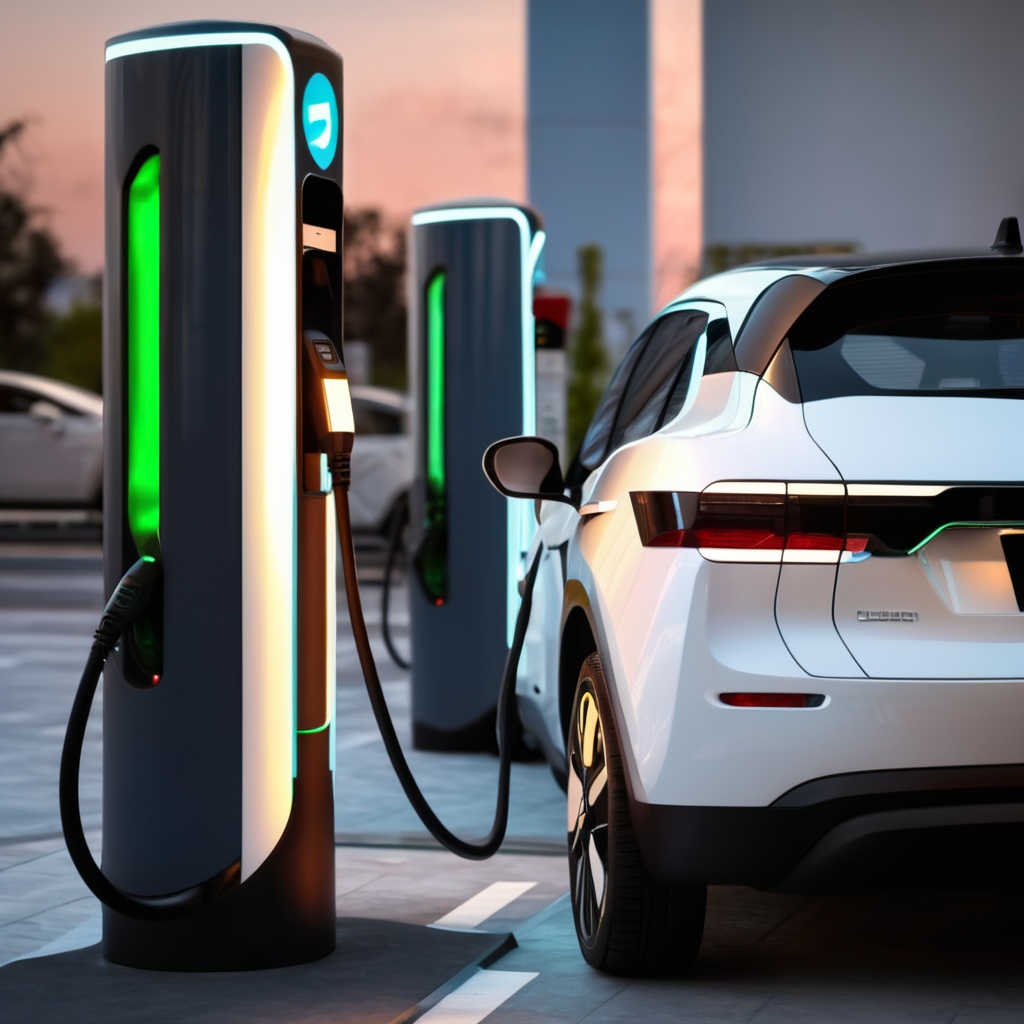} &
\includegraphics[width=\thumbw]{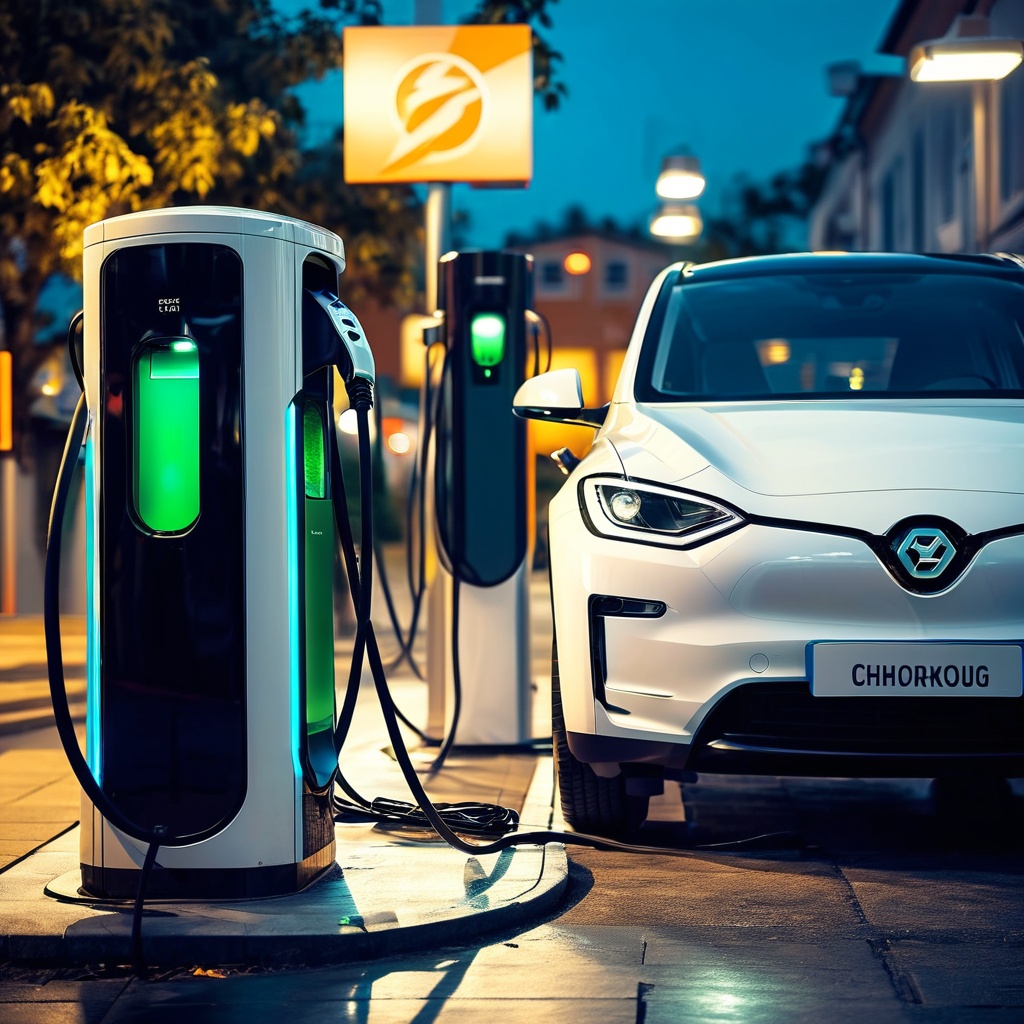} &
\includegraphics[width=\thumbw]{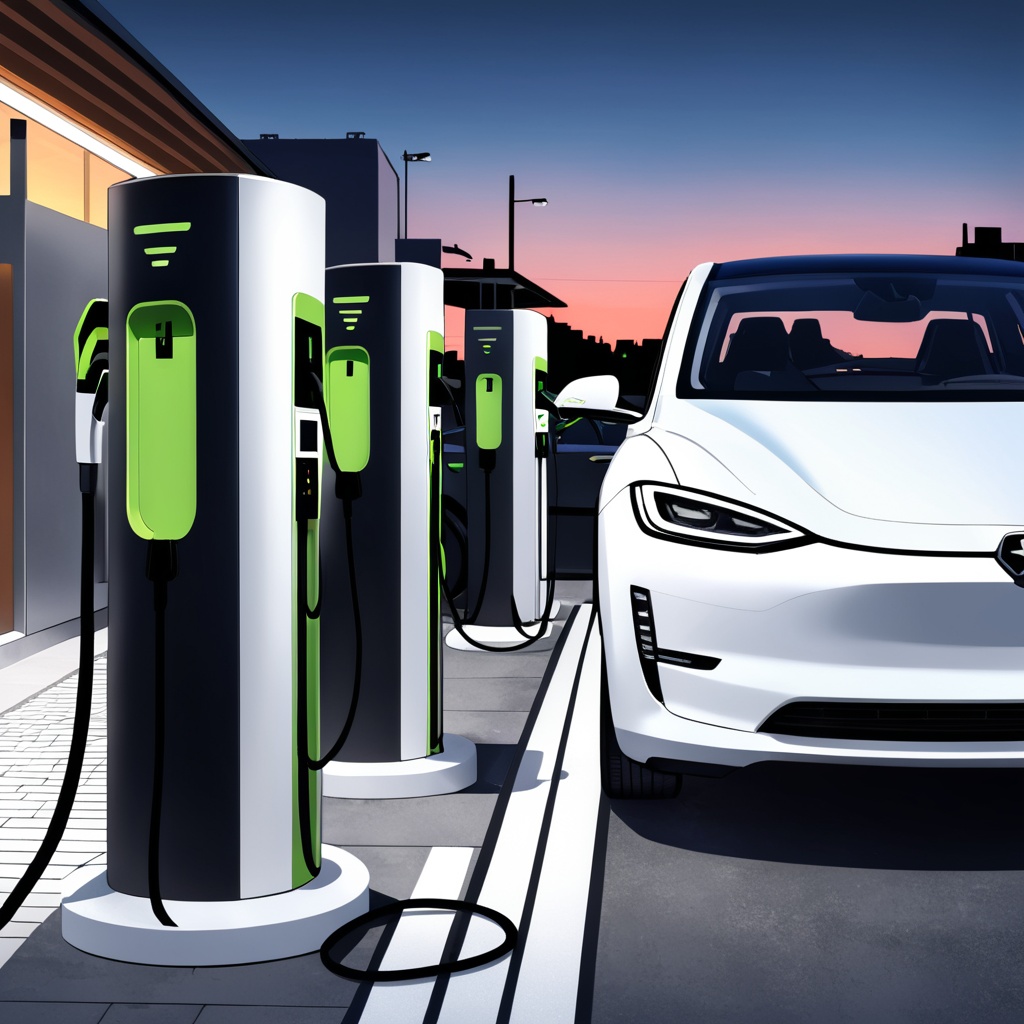} \\
\end{tabular}
\vspace{-2mm}
\caption{Sample images from our \textit{IntrAA-SD benchmark} dataset  generated using identical prompts by SDXL fine-tunes represented by M1: SDXL Base, M2: Juggernaut, M3: RealVis, M4: DreamShaper, M5: Animagine, M6: Counterfeit, M7: ZavyChroma, M8: Copax, M9: Proteus, M10: AAM Anime. P1: ``\emph{volcanic landscape with lava flow}'',
P2: ``\emph{electric car charging at a station}''. Despite sharing the same backbone, each fine-tune produces outputs with subtle visual distinctness with some pairs (like M2 \& M3) more similar than others. Note: We use M1-M10 to denote the corresponding fine-tuned models throughout the paper.} 
\vspace{-6mm}
\label{fig:intro_samples}
\end{figure*}
 

Therefore, ``intra-architecture" source attribution is required, where the target would be to identify fine-tunes of the same architectural base model, which is a fundamentally different task from architecture-level source attribution. An intra-architecture source attribution system should be designed to be capable of classifying images from differently fine-tuned models having the same architecture and pipeline, like the Juggernaut XL~\cite{juggernautxl} and RealVis XL~\cite{Yntec2024RealisticStockPhoto3} fine-tunes of SDXL.
But, as~\cite{bertazzini2025dragon, yang2022deepfake, sha2023fake} indicate, such an intra-architecture attribution using generic deep learning-based classifiers would be harder, as fine-tuning does not significantly alter the weights involved~\cite{ruiz2023dreambooth}, and obviously, the backbone remains the same.
Classifiers trained on images generated from such fine-tunes are prone to the influence of content and style, making the learning to distinguish between the subtle traces of the fine-tunes weak~\cite{deepfake2025}. When tested, they may fail, particularly when out-of-distribution prompts are used for generation, where the image style may no longer match those used during training~\cite{bertazzini2025dragon}.



In this paper, we introduce a novel task, \textit{IntrAA-SD}, intra-architecture source attribution for stable diffusion, where the source of the images generated from different fine-tunes of stable diffusion are to be identified. To this end, we compile a novel $140$K-image benchmark dataset containing images generated from $10$ different fine-tunes of the SDXL, along with in-domain (ID) and out-of-distribution (OOD) prompts. To the best of our knowledge, this is the first benchmark dataset for attribution within a single architecture family of fine-tuned models, and we focus on SDXL fine-tunes given their popularity. Fig.~\ref{fig:intro_samples} displays representative samples from our benchmark dataset, showing the subtle visual differences across different fine-tunes of a base model given identical prompts. We discuss the fine-tuned models used, elaborate the prompting strategies considered, and provide details about the image collection to form the dataset.

We also propose an initial baseline framework of intra-architecture attribution for the SDXL fine-tunes, which is designed to utilize a style-conditioned Mixture-of-Experts (SC-MoE) network. It uses a frozen CLIP backbone~\cite{radford2021learning} to capture image style that is used to gate a set of expert networks, where each expert processes model trace features of the fine-tunes from a LoRA~\cite{hu2021lora}-adapted DINOv2~\cite{oquab2023dinov2} operating at its native resolution. While DINOv2 is self-supervised with no use of text, making its features naturally suited for image structure representation~\cite{caron2021emerging}, LoRA refines the features for the source attribution task with the potential to emphasize the fine-tune traces. The gating of mixture-of-experts using style information from the CLIP backbone allows the trace features to be different for differently styled images. At the end, a classification is performed based on a concatenated feature representing both style and fine tune traces to achieve the intra-architecture source attribution. We rigorously evaluate the proposed baseline framework on our novel benchmark dataset in terms of attribution accuracy, and ablation studies are performed to investigate the effectiveness of its components.

\noindent To summarize, our key contributions are:
\begin{itemize}
\item We introduce a novel \textbf{Intr}a-\textbf{A}rchitecture \textbf{A}ttribution for \textbf{S}table \textbf{D}iffusion (\textbf{\textit{IntrAA-SD}}) task of identifying the specific fine-tuned model that generated a given image, among fine-tunes that share the same model architecture.
\item To facilitate the \textit{IntrAA-SD} task, we provide the novel \textbf{\textit{IntrAA-SD benchmark}} (ISDB) dataset of images generated using different fine-tunes of Stable Diffusion XL and various categories of prompts.
\item We propose an initial baseline, the \textbf{\textit{IntrAA-SD framework}}, utilizing style-conditioned MoE, CLIP and LoRA-adapted DINOv2 for performing the \textit{IntrAA-SD} task.
\end{itemize}

%% file: sec/2_formatting.tex
\vspace{-1mm}
\section{Related Work} \label{sec:review}
Although, to the best of our knowledge, there is no existing contribution specifically related to intra-architecture source attribution, we discuss several modern architecture-level source attribution techniques briefly below.

\begin{table*}[!t]
\centering
\caption{The Stable Diffusion XL (SDXL) fine-tunes used to generate our IntrAA-SD benchmark dataset.}
\label{tab:models}
\vspace{-2mm}
\begin{tabular}{ll}
\toprule
\textbf{Styles} & \textbf{Models} \\
\midrule
Photorealistic & SDXL Base 1.0~\cite{sdxl}, Juggernaut XL~\cite{juggernautxl}, RealVis XL~\cite{realvisxl}, DreamShaper XL~\cite{dreamshaperxl} \\
Anime & Animagine XL~\cite{animaginexl}, CounterfeitXL~\cite{counterfeitxl}, AAM XL Anime~\cite{aamxlanime} \\
Stylized & ZavyChromaXL~\cite{zavychromaxl}, CopaxTimelessXL~\cite{copaxtimelessxl}, ProteusXL~\cite{proteusxl} \\
\bottomrule
\end{tabular}
\vspace{-5mm}
\end{table*}

\subsection{Fake Image Detection and Attribution} 
Early work on fake image detection and source attribution includes detection of CNN-based generation, which shows that different architectures, datasets, and training methods leave detectable fingerprints in the generated images~\cite{wang2020}. A similar observation has been reported in the context of attributing images generated by generative adversarial networks (GANs), where the presence of rigid checkerboard artifacts created by the up-sampling layers (transposed convolutional layers)~\cite{frank2020leveraging} was leveraged.
Up-scaling operations alter the high-frequency spectrum of images~\cite{jain1989fundamentals} and generative models often produce spectral patterns that do not match real data~\cite{durall2020watch}, making the detection effective in the frequency domain. Thus, GAN based generation was found to be detectable by leveraging frequency domain spectral distortions~\cite{dzanic2020fourier}. 

Later, in~\cite{guarnera2024}, along with fake images from GANs, images generated from diffusion models were also identified and discriminated. A CLIP-based detector was used in~\cite{cozzolino2024raising} as it supports high generalizability, which is a key requirement for synthetic image detectors. Further, efforts have been made to minimize semantic bias by utilizing the frequency domain for the detection of diffusion-based image generations~\cite{ricker2022, bammey2024synthbuster}. It was demonstrated in~\cite{ricker2022} that unexpected convolutional traces remained in the high-frequency spectrum of images from diffusion models that use U-Net architectures.

\subsection{Source Attribution of Diffusion Models}

In the era of diffusion models, the focus has shifted from merely detecting fake images or attributing them to a broad class of generative models (like GAN, Diffusion, etc.), to examining the residual traces left by the diffusion process itself, enabling architecture-level source attribution~\cite{epstein2023online}. It was shown in~\cite{corvi2023} that diffusion models that uses different backbones (like Latent Diffusion, GLIDE, etc.) produce different spectral signatures based on the particulars of the noise scheduler and the up-sampler utilized in the generative process. 

The current leading methods for diffusion source attribution can be broadly organized into three categories, semantic analysis-based, spatial extraction-based and reconstruction-based. 

\textbf{Semantic Based:} Standard methods~\cite{clip_baseline, cozzolino2024raising} in this category primarily use frozen CLIP embeddings~\cite{clip_baseline} to identify and delineate semantic inconsistencies often found in synthetic images generated from different diffusion models. However, these methods have been found to be prone to overfitting, which in certain cases make CLIP-based classification detect styles rather than model fingerprints~\cite{bertazzini2025dragon}, making them ill-suited to detect fine-tuned sources.

\textbf{Spatial Based:} Methods in this category like DE-FAKE~\cite{sha2023fake}, use standard CNN backbones and cross-attention maps for diffusion source attribution and work well for classifying different generators such as Stable Diffusion, Midjourney, etc. These methods are also susceptible to false attribution when fine-tuned diffusion models are encountered, as they usually do not possess the required capacity to capture the subtle differences~\cite{bertazzini2025dragon}.

\textbf{Reconstruction Based:} 
This recent category of methods like the Diffusion Reconstruction Error (DIRE) method~\cite{wang2023dire} attempts diffusion model attribution by measuring the difference between the image at hand and its reconstructed counterpart, which is obtained by running forward diffusion using the image and then executing the pretrained reverse diffusion generation. A down-side to approaches like DIRE as an attribution technique is their resource-intensive nature, as rerun of the diffusion process is required. Use of DIRE for distinguishing images from fine-tuned diffusion models remains unexplored.



Until now, investigations on source attribution of diffusion models have been limited to architecture-level, like distinguishing Stable Diffusion (SD) v1.4 from Stable Diffusion (SD) v2.0. As mentioned earlier, such attribution techniques neither intend to distinguish fine-tunes of architecturally identical diffusion models nor many of them are suited for the same. Thus, the rapid growth of open-weight diffusion models naturally leads to the emergence of the intra-architecture source attribution problem, where the goal is to detect the nearly indistinguishable structural traces to delineate a fine-tune of the same architectural model from the other fine-tunes.

%% file: sec/3_finalcopy.tex
\vspace{-2mm}
\section{Novel Dataset: The IntrAA-SD Benchmark}
We build \textit{IntrAA-SD Benchmark} (ISDB) dataset in order to facilitate the attribution of a generated image to one of the fine-tunes of open-weight Stable Diffusion XL (SDXL) having the same architectural backbone, and to subsequently design such an intra-architecture source attribution baseline and evaluate it. We consider in-domain (ID) and out-of-distribution (OOD) prompts for the testing.

\begin{table*}[!t]
\centering
\caption{In-Domain (ID) prompt categories in our ISDB dataset with example prompts.}
\label{tab:id_prompts}
\vspace{-1mm}
\begin{tabular}{lll}
\toprule
\textbf{\#} & \textbf{Category} & \textbf{Example Prompt} \\
\midrule
\multicolumn{3}{l}{\textit{\textbf{Real-World Content (8 categories)}}} \\ \hline
1 & Portraits \& Faces & ``\emph{close up portrait of a young woman with freckles, natural light}'' \\
2 & Landscapes \& Nature & ``\emph{mountain range at sunrise with fog in the valley, golden light}'' \\
3 & Architecture \& Interiors & ``\emph{gothic cathedral interior with stained glass windows, light beams}'' \\
4 & Animals \& Wildlife & ``\emph{tiger walking through shallow water, jungle background, 8k detail}'' \\
5 & Food \& Objects & ``\emph{bowl of ramen with egg, nori, chopsticks, steam rising}'' \\
6 & Vehicles \& Scenes & ``\emph{red sports car on a mountain road, motion blur}'' \\
7 & Fantasy \& Sci-Fi & ``\emph{dragon flying over a medieval castle, fire breath, dark clouds}'' \\
8 & Abstract \& Artistic & ``\emph{oil painting of a stormy sea, thick impasto brushstrokes}'' \\
\midrule
\multicolumn{3}{l}{\textit{\textbf{Generative Stress Tests (7 categories)}}} \\ \hline
9 & Texture & ``\emph{woven fabric extreme close up, individual threads visible}'' \\
10 & Geometric \& Repetitive & ``\emph{infinite mirror tunnel, repeating reflections}'' \\
11 & Infeasible \& Indeterminate & ``\emph{Penrose triangle rendered as a real 3D object}'' \\
12 & Fluids \& Particles & ``\emph{milk drop splash crown forming, high speed macro}'' \\
13 & Lighting Extremes & ``\emph{single candle flame in complete darkness, warm glow}'' \\
14 & Text \& Signals & ``\emph{chalkboard filled with mathematical equations}'' \\
15 & Noisy \& Malicious & ``\emph{television static, random pixels, no coherent pattern}'' \\
\bottomrule
\end{tabular}
\end{table*}

\subsection{Generative Models Considered}
For intra-architecture source attribution, we generate images using $10$ SDXL fine-tunes from Hugging Face~\cite{wolf2020huggingfaces} spanning photorealistic, anime, and stylized categories as given in Table~\ref{tab:models}. This selection is done to include both challenging pairs in terms of attribution whose outputs are nearly indistinguishable to humans like Juggernaut~XL vs.\ RealVis~XL (refer to Fig.~\ref{fig:intro_samples}), and stylistically divergent pairs that primarily differ in visual style like Animagine~XL vs.\ RealVis~XL.

\subsection{In \& Out-of-Domain Prompting Strategies}

\begin{table*}[!t]
\centering
\caption{Out-of-Distribution (OOD) prompt categories in our ISDB dataset with example prompts.}
\label{tab:ood_prompts}
\vspace{-1mm}
\begin{tabular}{llp{12.2cm}}
\toprule
\textbf{\#} & \textbf{Category} & \textbf{Example Prompt} \\
\midrule
\multicolumn{3}{l}{\textit{Adversarial Strategies (7 categories)}} \\
\hline
1 & Cross-Domain & ``\emph{studio portrait, soft lighting, 85mm lens}'' $\rightarrow$ fed to Animagine XL; ``\emph{flat anime character, cel shading}'' $\rightarrow$ fed to RealVis XL \\
2 & Void Prompts & ``\emph{pure white background, nothing else''; ``complete darkness, empty frame}'' \\
3 & Nested Representation & ``\emph{a photograph of a charcoal sketch pinned to a cork board}''; ``\emph{a watercolor painting lying on a wooden desk}'' \\
4 & Text-Heavy & ``\emph{a handwritten letter on aged parchment, cursive text filling the page}'' \\
5 & Hands \& Anatomy & ``\emph{a pianist playing a grand piano, fingers on keys, concert hall}'' \\
6 & Reflections \& Glass & ``\emph{raindrops on a car window with blurred city lights, bokeh, night}'' \\
7 & High-Frequency Patterns & ``\emph{brick wall with every individual brick visible}''; ``\emph{woven fabric macro, individual threads}'' \\
\midrule
\multicolumn{3}{l}{\textit{Domain-Shift (11 categories)}} \\\hline
8 & Diverse Cultural & ``\emph{a colorful Indian Holi festival, powder clouds, joyful people}'' \\
9 & Medical \& Scientific & ``\emph{a microscope view of cells dividing, stained tissue sample}'' \\
10 & Historical \& Vintage & ``\emph{a daguerreotype style portrait, early photography aesthetic}'' \\
11 & Underwater \& Aquatic & ``\emph{a diver exploring a sunken shipwreck, flashlight beam, deep blue}'' \\
12 & Sports \& Action & ``\emph{a basketball player dunking, arena lights, freeze frame}'' \\
13 & Fashion \& Textile & ``\emph{haute couture dress on a runway, dramatic lighting}'' \\
14 & Mythology \& Folklore & ``\emph{a Japanese kitsune spirit in a misty bamboo forest}'' \\
15 & Industrial \& Decay & ``\emph{a derelict shopping mall, broken escalators, overgrown plants}'' \\
16 & Unusual Perspectives & ``\emph{fisheye lens view of a skate park, extreme wide angle distortion}'' \\
17 & Weather \& Atmospheric & ``\emph{a tornado funnel cloud touching down on farmland, debris}'' \\
18 & Edge Cases & ``\emph{a completely white image with a barely visible gray shape}'' \\
\midrule
\multicolumn{3}{l}{\textit{Subtle Domain-Shift (2 categories)}} \\ \hline
19 & Lighting Condition & ``\emph{harsh midday sun, deep black shadows on white architecture}'' \\
20 & Surreal \& Impossible & ``\emph{an impossible staircase like MC Escher, paradox geometry}'' \\
\bottomrule
\end{tabular}
\end{table*}

\begin{figure*}[t]
    \centering
    \includegraphics[width=0.9\textwidth, page=1]{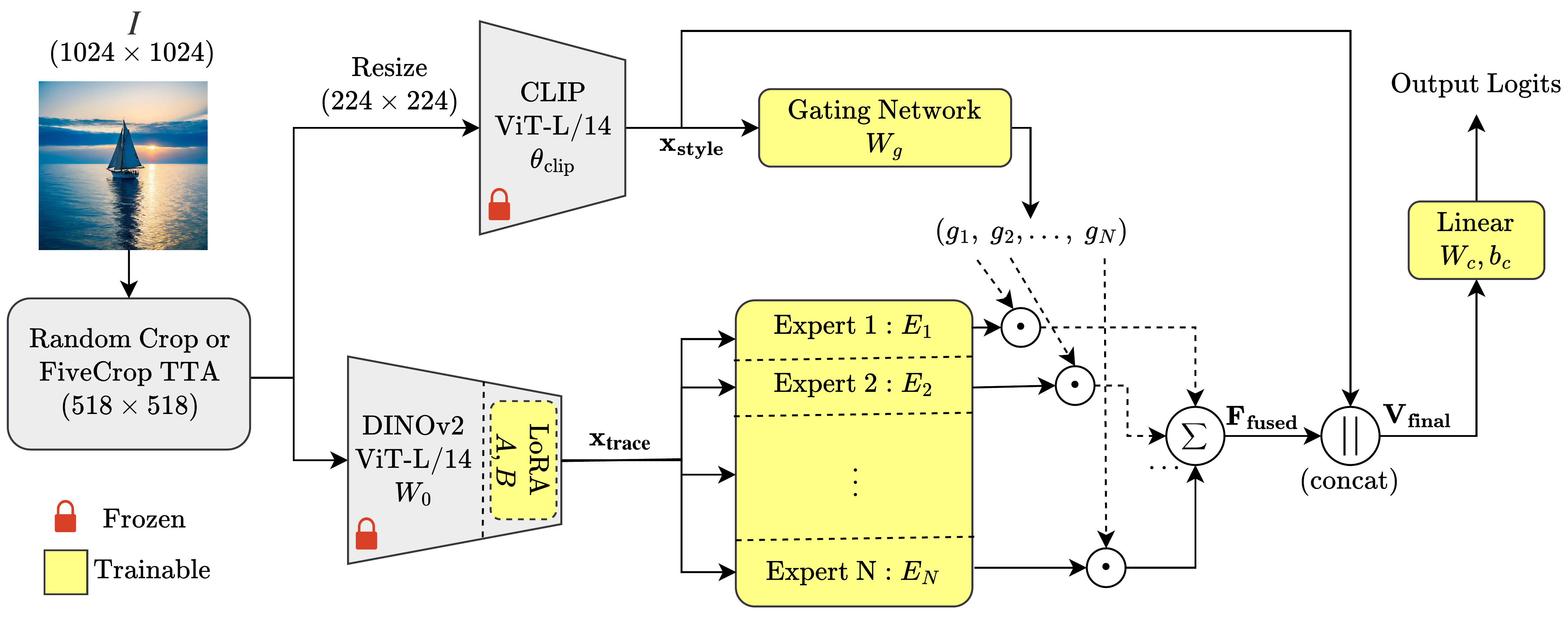}
    \vspace{-2mm}
    \caption{Our IntrAA-SD Baseline Framework. A frozen CLIP backbone captures the image style (Sec.~\ref{sec:clip_extraction}) to compute gating weights across $N$ experts. A LoRA-adapted DINOv2 backbone extracts the fine-tunes' related patterns fed into experts (Sec.~\ref{sec:dinov2_extraction}). The expert outputs are combined via style-conditioned Mixture-of-Experts (Sec.~\ref{sec:sc_moe}) and concatenated with the style context for the attribution (Sec.~\ref{sec:fusion}).} 
    \label{fig:architecture}
    \vspace{-6mm}
\end{figure*}

We created two prompt sets, in-domain (ID) and out-of-distribution (OOD) for testing, and different prompts from the ID set are used to train the SDXL fine-tunes.

The ID set contains 300 prompts spread across 15 categories as shown in Table~\ref{tab:id_prompts}. Eight of these categories cover standard real-world  content such as portraits, landscapes, architecture, animals and food. The remaining seven are particularly challenging in terms of generation, which involves image generation with repetitive textures and geometric patterns, extreme lighting, noise, etc.~\cite{corvi2023}. Some such categories under the type of generative stress test listed in Table~\ref{tab:id_prompts} are elaborated in the \textit{supplementary}.

The OOD set contains 200 prompts across 20 categories as listed in Table~\ref{tab:ood_prompts}. These categories fall under adversarial strategies and domain shift ones. In the latter, a couple of categories represent subtle domain shift, where the prompts are similar to a few in the ID set but demand quite different generations.
Adversarial strategies contain categories such as cross-domain prompting, void prompts, nested representations, etc. Due to its adversarial nature, the cross-domain prompting category is mentioned here and not under domain shift strategy. In cross-domain prompting category, photorealistic prompts are fed to anime-focused models like Animagine XL, and conversely, cartoon and illustration prompts to photorealistic models like RealVis XL. This makes it difficult for a intra-architecture source attribution technique to classify, as the generated images could be quite different than those on which the attribution model is trained. 

A couple more categories under adversarial strategies listed in Table~\ref{tab:ood_prompts} are elaborated in the \textit{supplementary}.
The other OOD prompt categories like cultural scenes, medical and scientific imagery, historical styles, underwater environments, sports and fashion are mentioned under domain-shift, as they are in general not adversarial in nature. Their use to generate images different from those generated using ID prompts is straightforward.

\subsection{Dataset Details}
Our \textit{ISDB} dataset consists of $140$K images, evenly split across $10$ classes ($12$K ID \& $2$K OOD images per class),  with each class representing generation from an SDXL fine-tune. Every image is generated with $1024 \times 1024$ resolution using the Hugging Face Diffusers library~\cite{von-platen-etal-2022-diffusers}. We have used the DPM{\small ++} SDE Karras scheduler~\cite{lu2022dpm-solver,karras2022elucidating} with 25 denoising steps, a classifier-free guidance (CFG) scale of 7.0~\cite{ho2022classifier}, and a fixed negative prompt to suppress common artifacts.  All outputs are saved at 95\% JPEG quality to simulate real-world dissemination conditions, since most images encountered in the wild have gone through at least one round of lossy compression~\cite{wang2020}. The ID set comprises $300 \textrm{ (prompts)} \times 40 \textrm{ (seeds)}= 12000$ images per model of 15 prompt categories, while the OOD set adds $200 \textrm{ (prompts)} \times 10 \textrm{ (seeds)}= 2000$ images per model from 20 prompt categories. We split the ID set into 80/20 train and test partitions using stratified sampling with a fixed random state, ensuring no prompt overlap between splits. The OOD set is entirely used only for evaluation. ID and OOD set images are rendered with 40 and 10 deterministic seeds (to random number generator), respectively, to ensure diversity while maintaining reproducibility.
    

\vspace{-2mm}
\section{Initial Intra-architecture Source Attribution Baseline: The IntrAA-SD Framework}
\vspace{-1mm}
We propose the first baseline, our IntrAA-SD framework, for intra-architecture source attribution of Stable Diffusion XL (SDXL) fine-tunes. The key idea in our framework is to decouple the style understanding from the fine-tune trace analysis, and utilize both judiciously. As illustrated in Fig.~\ref{fig:architecture}, the framework has four components, a frozen CLIP backbone~\cite{radford2021learning} that captures the visual style of the input, a LoRA-adapted DINOv2 backbone~\cite{oquab2023dinov2} that extracts fine-tune trace related features at its native pixel resolution, a style-conditioned Mixture-of-Experts (MoE)~\cite{fedus2022switch} where CLIP-derived style information gates the trace features through specialized experts, and finally, a classifier that acts on a fused feature content to give the final logits. Our framework takes generated images as inputs at $518 \times 518$ resolution, with random crop employed during training and five crop test-time augmentation during testing.
\vspace{-3mm}
\subsection{Visual Style Extraction}\label{sec:clip_extraction}
\vspace{-1mm}
A common aspect that may hinder intra-architecture attribution is that classifiers would learn the predominant image style rather than the subtle traces left by the generator~\cite{bertazzini2025dragon}. For instance, a model trained on our dataset might learn to associate anime-style output with Animagine XL, as the training images generated by it would mostly be anime. As evident from Table~\ref{tab:models}, while style can help distinguish some fine-tunes from others, it alone may not be sufficient to pinpoint a fine-tuned generator in general. Dependence on style alone could result in failure, particularly in OOD cases such as when we feed a photorealistic prompt to Animagine XL. To address this, we intend to ensure that subtle fine-tune traces are leveraged along with style for intra-architecture source attribution. 
\par For extracting the style features ($\mathbf{x}_{\text{style}}$) from a given image, we use a frozen CLIP ViT-L/14~\cite{radford2021learning}. As CLIP was trained on a large-scale dataset of image-text pairs, it has learned rich representations of visual style and composition~\cite{radford2021learning}. So, it would capture the same from the input images. More details on this are shared in the \textit{supplementary}.
\vspace{-1mm}
\subsection{Fine-tune Trace Feature Extraction}
\label{sec:dinov2_extraction}
\vspace{-1mm}

For extracting features representing fine-tune traces in image structure, we choose a frozen DINOv2 ViT-L/14~\cite{oquab2023dinov2} followed by a trainable Low-Rank Adaptation (LoRA)~\cite{hu2021lora}. Unlike CLIP, DINOv2 is self-supervised and do not involve text. So, the pretrained DINOv2 has learned rich representations of visual structure, texture and content~\cite{caron2021emerging}. Working at its native high resolution~\cite{oquab2023dinov2}, DINOv2 has the capability to represent image details~\cite{ricker2022, corvi2023}. Then, to consider only the features relevant to intra-architecture source attribution using our framework, trainable LoRA is adopted. LoRA therefore emphasizes only fine-tuned generator relevant features distilling them from the image structure and content representations in its low-rank representation.
We represent the fine-tune trace features obtain from LoRA-adapted DINOv2 as a vector $\mathbf{x}_{\text{trace}} \in \mathbb{R}^{d_{\text{trace}}}$, where $d_{\text{trace}} = 1024$.

\vspace{-1mm}
\subsection{Style-Conditioned Mixture of Experts}\label{sec:sc_moe}
\vspace{-1mm}
Differently styled generated images may contain structurally different generator signatures~\cite{wang2020,frank2020leveraging}. Therefore, rather than relying on a single classifier to implicitly handle this variation, we consider a style-conditioned gating driven Mixture-of-Experts (SC-MoE) architecture before classification.
We adopt an MoE architecture~\cite{shazeer2017outrageously, fedus2022switch} with a cross-modal gating
of model trace features from LoRA-adapted DINOv2. The gating decision, which dictates how much each expert contributes to the final output, is determined by a gating network working on the style feature vector $\mathbf{x}_{\text{style}}$ from CLIP. 
The gating weights are computed in the gating network as 
$g = \text{Softmax}(W_g \mathbf{x}_{\text{style}})$, where  $g \in \mathbb{R}^{N}$~\cite{shazeer2017outrageously} with $N$ being the number of experts.
Every expert $E_i$ is a two-layer MLP with output dimension,
$d_{\text{expert}}{=}1024$, and their outputs are combined using the weights, $g_i$, to get the fused vector $\mathbf{F}_{\text{fused}}\in \mathbb{R}^{d_{\text{expert}}}$.
\subsection{Classification for Intra-architecture Source Attribution}\label{sec:fusion}
To leverage both the fine-tune trace and style information from the image, we concatenate the clearly separated SC-MoE fused trace features $\mathbf{F}_{\text{fused}}$ and style features $\mathbf{x}_{\text{style}}$ 
to form a joint representation $\mathbf{V}_{\text{final}}$ that carries both fine-tuned model signature and image style context.

Specifically, the fused SC-MoE output $\mathbf{F}_{\text{fused}} \in \mathbb{R}^{d_{\text{expert}}}$ ($d_{\text{expert}} = 1024$) is concatenated with the CLIP style vector $\mathbf{x}_{\text{style}} \in \mathbb{R}^{d_{\text{style}}}$ ($d_{\text{style}} = 1024$) as:
\begin{equation}
\nonumber
    \mathbf{V}_{\text{final}} = [ \mathbf{F}_{\text{fused}} \parallel
    \mathbf{x}_{\text{style}} ] \in
    \mathbb{R}^{d_{\text{expert}} + d_{\text{style}}} 
    (\mathbb{R}^{2048}) .
\end{equation}
This 2048-dimensional vector is then passed through a linear layer with the weight matrix $W_c \in \mathbb{R}^{C \times 2048}$ and bias $b_c \in \mathbb{R}^{C}$. The training of the entire system is done predominantly using the standard cross-entropy loss $\mathcal{L}_{CE}$ over $C$ classes (number of fine-tuned generators considered).
The overall training objective is:
\begin{equation}
\label{Eq1}
    \mathcal{L}_{\text{total}} = \mathcal{L}_{CE}
    + \lambda_{\text{bal}} \mathcal{L}_{\text{bal}}, 
\end{equation}
where $\lambda_{bal}$ is a hyperparameter to adjust the weightage of load-balancing loss, $\mathcal{L}_{\text{bal}}$, 
which is designed to make the model use all the experts in the MoE more equally~\cite{shazeer2017outrageously}. When our \textit{ISDB} dataset is considered, we have $C=10$. Thus, by providing the classifier both the model trace information and the style context, we enable it to learn generator separation based on different model trace signatures for different styles.
\begin{table}[!t]
\centering
\caption{Intra-architecture source attribution accuracies (\%) of our \textit{IntrAA-SD} baseline framework on our ISDB dataset, for the test images generated with In-Domain (ID) and Out-of-Distribution (OOD) prompts.}
\label{tab:main_results}
\vspace{-3mm}
\begin{tabular}{@{}l|cc@{}}
\toprule

\textbf{Source Model} & \textbf{ID} & \textbf{OOD} \\
\midrule 
SDXL Base  (M1)      & {99.42} & {91.50} \\
Juggernaut XL  (M2) & {99.79} & {70.50} \\
RealVis XL    (M3)   & {98.67} & {82.50} \\
DreamShaper XL (M4)  & {99.83} & {99.00} \\
Animagine XL   (M5) & {99.92} & {98.00} \\
CounterfeitXL  (M6)  & {98.29} & {98.00} \\
ZavyChromaXL   (M7)  & {99.67} & {74.00} \\
CopaxTimelessXL (M8) & {97.88} & {90.00} \\
ProteusXL    (M9)  & {98.58} & {90.50} \\
AAM XL Anime  (M10) & {98.21} & {91.50} \\
\midrule
\textbf{Overall}  & \textbf{99.03} & \textbf{88.55} \\
\bottomrule
\end{tabular}%
\vspace{-6mm}
\end{table}

\vspace{-2mm}
\section{Experiments}
\vspace{-1mm}
As we have introduced the new task of intra-architecture source attribution in this paper with a novel benchmark dataset and a baseline technique, we evaluate the baseline technique on the ISDB dataset in this section and perform its empirical ablation study. Due to space constraints here, we report additional experiments, and investigate the applicability of existing architecture-level attribution models for intra-architecture attribution in the \textit{supplementary}.

\vspace{-1mm}
\subsection{Training Strategy}
\vspace{-1mm}
The \textit{IntrAA-SD} baseline framework is trained for 30 epochs on the in-domain (ID) training set of our ISDB dataset employing AdamW~\cite{loshchilov2017decoupled} and cosine annealing with warm restarts~\cite{loshchilov2017sgdr}. The LoRA parameters use a learning rate of $1 \times 10^{-4}$, while the MoE and classification heads use $5 \times 10^{-4}$. We also use label smoothing ($\epsilon = 0.1$)~\cite{labelsmoothing}. For augmentation, we apply MixUp ($\alpha=0.2$)~\cite{mixup} and CutMix ($\alpha=1.0$)~\cite{yun2019cutmix} with  a probability $p=0.5$. Random cropping is used for training, and FiveCrop Test Time Augmentation is applied at inference.

\subsection{Attribution Results for ID and OOD Prompts}



Table~\ref{tab:main_results} presents the results of our \textit{IntrAA-SD} framework considering images generated in our ISDB dataset from all the $10$ fine-tuned source models using in-domain (ID) and out-of-distribution (OOD) testing prompts, showing the per-class (source) accuracies. 
Our method achieves an average of 99.03\% accuracy on the ID test set case and 88.55\% on OOD case. We refer to the results in Table~\ref{tab:main_results} further in the \textit{supplementary} for more discussion.

\subsection{Ablation Studies on \textit{\textbf{IntrAA-SD}} framework}
\subsubsection{Core Modules}
In Table~\ref{tab:component_ablation}, we report an ablation study of our baseline intra-architecture attribution method using our ISDB dataset's ID test set, to note the significance of its components.

\textbf{Gating strategy.}
As seen in Table~\ref{tab:component_ablation}, the equal weighting case (A1), where the gating network is entirely removed and equal-weighted average of all the experts is considered, yields 98.40\% accuracy. The DINOv2 driven variant (A2), where features from frozen DINOv2 (not trained with text) are used in the gating network to determine routing weights, produces almost a similar accuracy of 98.42\%.
However, when use consider the use of style features from frozen CLIP in our \textit{IntrAA-SD} framework to drive the gating network, the highest accuracy of 99.03\% is achieved.
This observation demonstrates that marginal benefit is achieved by specifically designing the extraction of style features and using it for gating the Mixture-of-Experts (MoE) network.

\begin{table}[!t]
\centering
\caption{\textit{IntrAA-SD} Ablation: significance of components. All variants use LoRA r=16, 518px, 4 experts (unless removed).}
\label{tab:component_ablation}

\begin{tabular}{p{1.7cm}|p{4.5cm}|l}
\hline
\textbf{Components} & \textbf{Variants} & \textbf{Acc \%}  \\
\hline
 \multirow{3}{*}{Gating } 
  & A1: Equal weighting (no gating) & \multirow{2}{*}{98.40 } \\
 \multirow{1}{*}{strategy} & A2: Structure-based gating (DINOv2 $\rightarrow$ gate) & \multirow{2}{*}{98.42 } \\

\hline
 \multirow{2}{*}{Style concat}& {A3: No style concatenation (after MoE) before classification} & \multirow{2}{*}{98.21}   \\
\hline
\multirow{3}{*}{Style or}
  & A4: Image Style-only (Frozen CLIP + linear head) & \multirow{2}{*}{87.59}  \\
 \multirow{1}{*}{trace feature}  & A5: Source trace-only (LoRA DINOv2 + linear head) & \multirow{2}{*}{98.52} \\
\hline
All & \textit{IntrAA-SD} framework & \textbf{99.03} \\
\hline
\end{tabular}
\vspace{-6mm}
\end{table}

\begin{table*}[!t]
\centering
\caption{Ablation on Hyperparameters. Ours: \textit{IntrAA-SD} baseline intra-architecture source attribution framework} 
\label{tab:arch_ablation}
\small
\vspace{-2mm}
\begin{tabular}{c |c|c || c | c|c}
\hline
& \textbf{Variants} & \textbf{Acc (\%)} & & \textbf{Variants} & \textbf{Acc (\%)} \\
\hline
\multirow{3}{*}{Model trace extractor}
  & Frozen DINOv2        & 93.01 &
\multirow{3}{*}{Input resolution}
  & $224\times224$          & 96.75 \\
  & LoRA $r{=}4$         & 98.50 &
  & $378\times378$          & 98.41 \\
  & LoRA $r{=}16$ (Ours) & \textbf{99.03} &
  & $518\times518$ (Ours)   & \textbf{99.03} \\
\hline \hline
\multirow{4}{*}{Number of experts}
  & 1 (no MoE)    & 98.77 &
\multirow{4}{*}{Inference input}
&& \\
& 2             & 98.84 &
  &  Center Crop         & 98.23\\
  & 4  (Ours)     & \textbf{99.03} & &Five Crop TTA (Ours)   & \textbf{99.03}  \\
  & 8              & 98.88 & & & \\
\hline
\end{tabular}
\vspace{-4mm}
\end{table*}

\textbf{Style feature concatenation.}
From Table~\ref{tab:component_ablation}, we infer that the concatenation of style features $\mathbf{x}_{\text{style}}$ with fused fine-tuned model trace features $\mathbf{F}_{\text{fused}}$ from the MoE helps our \textit{IntrAA-SD} framework to achieve 99.03\% accuracy, while if the concatenation is not performed (A3) and only $\mathbf{F}_{\text{fused}}$ is fed into the classifier, the accuracy drops to 98.21\%.


\textbf{Style \& source trace.}
We attempt to classify the generated input images for detecting the fine-tuned source model using only the style features from the frozen CLIP without using any MoE network (A4). This yields a poor 87.59\% accuracy as seen in Table~\ref{tab:component_ablation}. This is expected, as from our discussion in Section~\ref{sec:clip_extraction}, we know that style features may not be sufficient on its own for intra-architecture attribution. In contrast, when we attempt the same using only the fine-tune trace features from the frozen DINOv2 and LoRA removing the MoE network (A5), a much improved 98.52\% accuracy is obtained. This shows that the frozen DINOv2 with LoRA indeed provides features relevant to intra-architecture attribution. When both the source trace features to drive the MoE and the style features primarily to gate the MoE are used in our \textit{IntrAA-SD} framework, we achieve the best performance with an accuracy of 99.03\%.
\subsubsection{Hyperparameter Choices}
We also study the impact of hyperparameter and inference choices, and the related performance is listed in Table~\ref{tab:arch_ablation}.

\textbf{Fine-tune model trace feature extractor.}
In Table~\ref{tab:arch_ablation}, we see that if a completely frozen DINOv2 alone is used to extract intra-architecture attribution relevant features, only 93.01\% accuracy is obtained. However, injecting trainable LoRA matrices improves the performance substantially achieving 98.5\% and 99.03\% accuracies for ranks (r) 4 and 16, respectively. This suggests that LoRA indeed effectively distills model trace features from the image structure represented by DINOv2 as discussed in Section~\ref{sec:dinov2_extraction}.


\textbf{Number of experts in MoE.}
We experimented by changing the number of experts in MoE in our framework. As seen from Table~\ref{tab:arch_ablation}, a slight increase in accuracy from 98.77\% to 99.03\% is observed when four experts are used instead of one, which represents the case of no MoE. Although the effect of MoE on the performance may not be that significant, we stick with the four experts case to keep the fraction of gain attained.

\textbf{Resolution and inference.}  In Table~\ref{tab:arch_ablation}, we see that the use of $518\times518$ resolution inputs gives the 
best accuracy of 99.03\% among the listed resolutions. The higher accuracy at higher resolution is expected as model trace signatures embedded in spatial details would be more accurately represented. Further, we see that five crop test-time augmentation (TTA) provides an improved accuracy over single center crop during inference.

\textbf{Load-balancing loss weightage.} In Table~\ref{tab:supp_lambda}, when a load-balancing loss is employed in the loss function of Equation (\ref{Eq1}) with a small weight of $.001$, the best accuracy of 99.03\% is attained, and hence, it is used in our baseline.


\begin{table}[t]
\centering
\caption{Ablation on the loss function ($\mathcal{L}_{total}$) parameter $\lambda_{bal}$}
\label{tab:supp_lambda}
\begin{tabular}{c|c|c|c|c}
\hline
\textbf{$\lambda_{bal}$} & 0.0 (No balancing) & \textbf{0.001} & 0.01 & 0.05 \\
\hline
\textbf{Acc (\%)} & 98.78 & \textbf{99.03} & 98.72 & 98.70 \\
\hline
\end{tabular}
\vspace{-4mm}
\end{table}

\subsection{Fine-tunes' Trace Feature Space Visualization}
\begin{figure}[!t]
    \centering
    \includegraphics[width=0.95\linewidth]{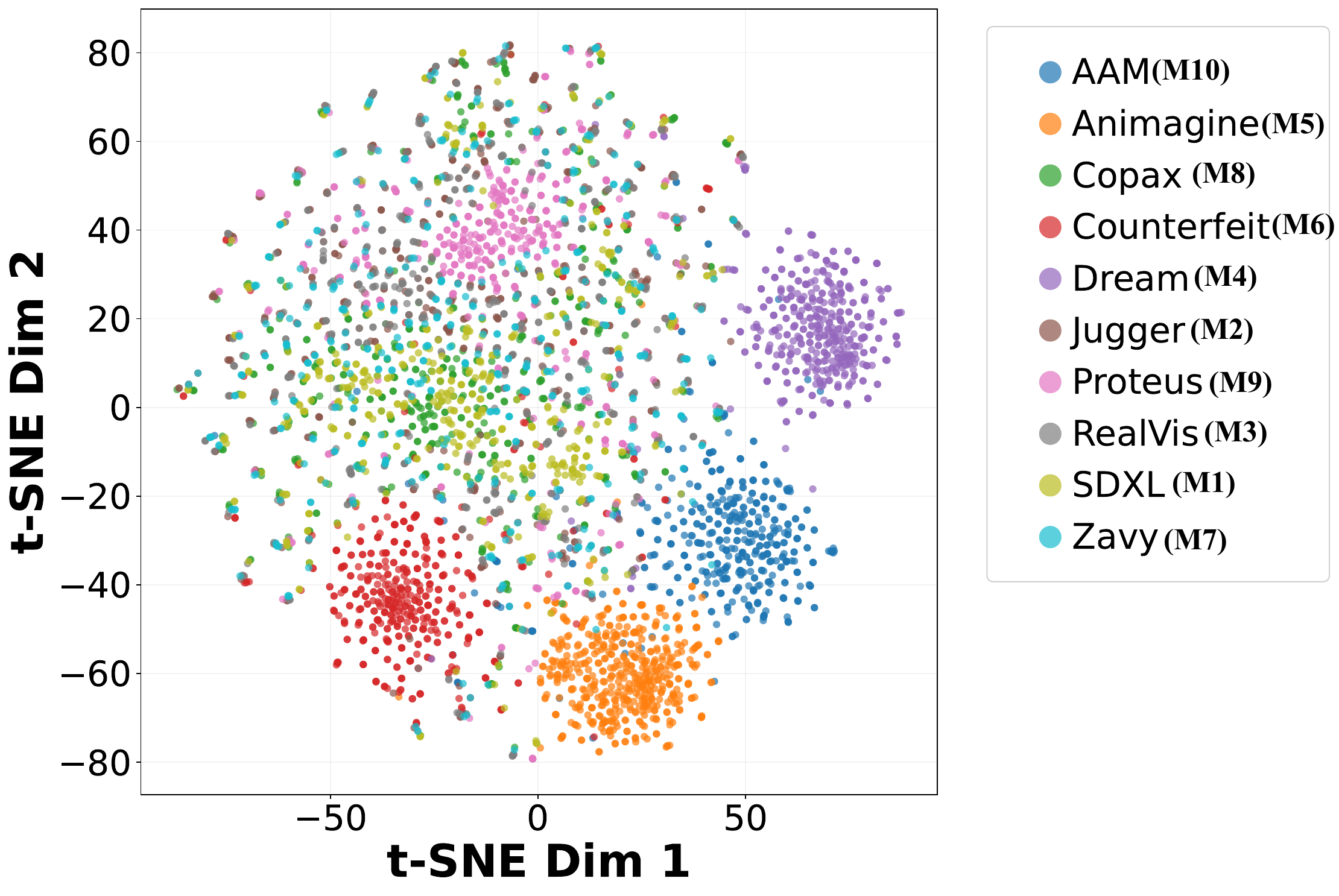}
    \caption{t-SNE visualization of model trace feature embeddings.}
    \label{fig:tsne}
    \vspace{-4mm}
    \end{figure}

We project the 1024-d $\mathbf{F}_{\text{fused}}$ model trace features onto two dimensions using t-SNE, as shown in Fig.~\ref{fig:tsne}. 
As can be seen, images from the fine-tuned SDXL models such as Animagine, Counterfeit, AAM and DreamShaper are represented as compact clusters well-separated from the others. Some fine-tunes like Proteus appear reasonably separable, but the few others do not appear so in the 2-d plot. However, as we have seen earlier, the fine-tune trace feature representation in our baseline model actually achieves more than 99\% source attribution accuracy, which suggests that most of the fine-tunes must be reasonably well-separated in the $1024$ dimensional feature space.

\section{Conclusion}
\vspace{-2mm}
Identifying the precise source of images produced by fine-tuned generators, which share the same architectural backbone with other fine-tunes, is non-trivial and essentially a separate task than the existing architecture-level source attribution. In this regard, we introduced a novel task, \textit{IntrAA-SD}, Intra-Architecture source Attribution for Stable Diffusion, provided a first-of-its-kind $140$K-image benchmark dataset (IntrAA-SD Benchmark - ISDB), and proposed the initial baseline framework (IntrAA-SD framework) to perform the Intra-architecture attribution of fine-tunes. Our ISDB dataset is formed considering $10$ Stable Diffusion XL fine-tunes, and both in-domain (ID) and out-of-distribution (OOD) prompts. Our baseline approach leverages style-conditioned Mixture-of-Experts (MoE), pretrained CLIP and LoRA-adapted DINOv2.

The baseline IntrAA-SD framework is trained and tested on our ISDB dataset. Elaborate ablation studies are also performed on the framework components, hyperparameters and the loss function, which reveal the significance of certain design choices. While our baseline approach achieves a high attribution accuracy of 99.03\% accuracy for the ID test images of our dataset, it does reasonably well with 88.55\% accuracy for the OOD test images. These performances are reasonably higher than those achieved by training the existing architecture-level source attribution models in our ISDB dataset, as those models are not designed for intra-architecture source attribution. These findings suggest that there is immense scope for future investigations targeting development of intra-architecture source attribution models and compilation of more relevant datasets, picking up from our contribution. 